\documentclass{article}

\usepackage[preprint]{corl_2022} 

\usepackage{graphicx}
\usepackage{amsmath}
\usepackage{mathtools, nccmath}
\usepackage{lipsum}
\usepackage{physics}
\usepackage{adjustbox}
\usepackage{amsfonts}
\title{Toward a Framework for Adaptive Impedance Control of an Upper-limb Prosthesis}
\author{
  Laura Ferrante\\
  \small{School of Computer Science,}\\
  \small{University of Birmingham,}\\
  \small{Birmingham, UK}\\
  \small{\texttt{lxf656@student.bham.ac.uk}} \\
  \And
  Mohan Sridharan\\
  \small{School of Computer Science,}\\
  \small{University of Birmingham,}\\
  \small{Birmingham, UK}\\
  \small{\texttt{M.Sridharan@bham.ac.uk}} \\
  \And
  Claudio Zito\\
  \small{Autonomous Robotics Research Centre,}\\
  \small{Technology Innovation Institute,}\\
  \small{Abu Dhabi, UAE}\\
  \small{\texttt{Claudio.Zito@tii.ae}} \\
\And
  Dario Farina\\
  \small{Department of Bioengineering}\\
  \small{Imperial College London,}\\
  \small{London, UK}\\
  \small{\texttt{D.Farina@imperial.ac.uk}}\\
}

\begin{document}
\maketitle
\begin{abstract}
Adapting upper-limb impedance (i.e., stiffness, damping, inertia) is essential for humans interacting with dynamic environments for executing grasping or manipulation tasks.
On the other hand, control methods designed for state-of-the-art upper-limb prostheses infer motor intent from surface electromyography (sEMG) signals in terms of joint kinematics, but they fail to infer and use the underlying impedance properties of the limb.
We present a framework that allows a human user to simultaneously control the kinematics, stiffness, and damping of a simulated robot through wrist's flexion-extension. 
The framework includes muscle-tendon units and a forward dynamics block to estimate the motor intent from sEMG signals, and a variable impedance controller that implements the estimated intent on the robot, allowing the user to adapt the robot's kinematics and dynamics online. 
We evaluate our framework with 8 able-bodied subjects and an amputee during reaching tasks performed in free space, and in the presence of unexpected external perturbations that require adaptation of the wrist impedance to ensure stable interaction with the environment. 
We experimentally demonstrate that our approach outperforms a data-driven baseline in terms of its ability to adapt to external perturbations, overall controllability, and feedback from participants. 
\end{abstract}




\maketitle
\begin{figure*}
\centering
\includegraphics[width=1\textwidth]{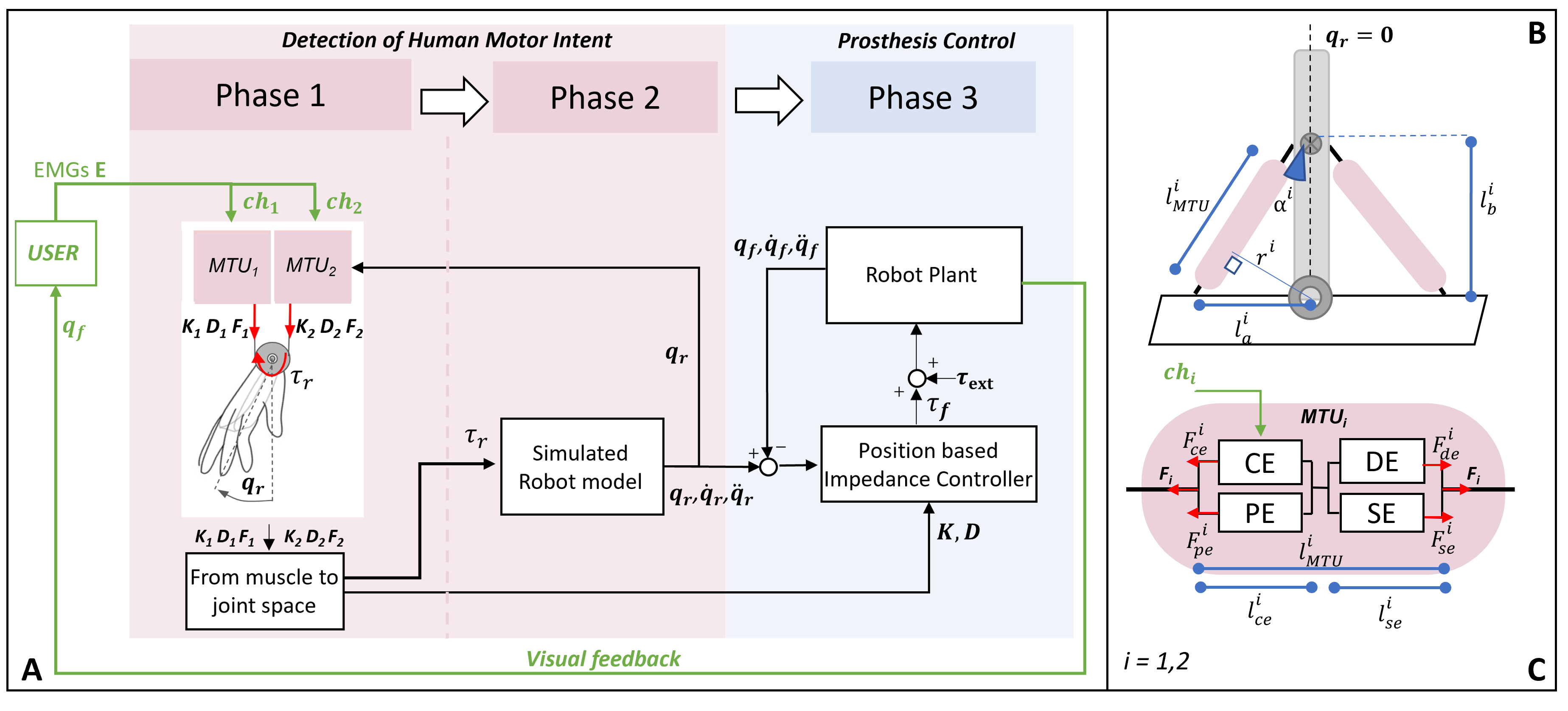}
\caption{\textbf{(A)} Overview of our framework to control a 1-DoF robot. It comprises a \textit{Detection of Human Motor Intent} block and a \textit{Prosthesis Control} block. The first block maps sEMG signals ($ch_{1}, ch_{2}$) from the user's forearm muscles to estimates of the user's motor intent in terms of kinematics, i.e., $s_{r} = (q_{r},\dot{q}_{r},\ddot{q}_{r})$, and dynamics, i.e., joint stiffness $K$ and damping $D$. The second block executes the motor intent on the robot plant, using a variable impedance controller to track $s_{r}$ based on $K$ and $D$. The framework's output $q_{f}$ is also used as visual feedback by the human user to modulate impedance in response to external perturbation on the robot. \textbf{(B)} Arrangement of the $MTUs$ on the link of the simulated robot model. \textbf{(C)} Forces generated by the $MTU$'s elements; $MTU_{i}$ comprises muscle (CE, PE) and tendon components (DE, SEE) of length $l_{ce}^{i}$ and $l_{se}^{i}$ respectively.}
\label{fig:framework}
\end{figure*}

\section{Introduction}
\label{sec:intro}
Consider a human carrying a cup of coffee; forces arise from the interaction between the liquid, the cup, and the human hand. To ensure stable interaction with the cup and external perturbations, humans modulate their limb impedance, i.e., stiffness, damping, and inertia, by co-activating agonist and antagonist muscles \cite{hogan1984adaptive}. 
Modulation of the impedance properties of the robot (i.e., the prosthesis) is thus a crucial aspect in prostheses control given that the user with prosthetic limbs physically interacts with the environment.
However, none of the commercially available upper-limb prostheses allow the user to modulate impedance properties because it is challenging to accurately decode human motor intent from low-bandwidth surface electromyography (sEMG) signals~\cite{hahne2017user}, and due to the high redundancy of the human motor control. In fact, the mapping from sEMG signals to target joint kinematics is not unique since the same action can be performed at different levels of contraction of muscles.

Data-driven methods, which are the state of the art in prosthesis control of commercially available devices, map sEMG signals to control commands (i.e., kinematics) in an offline training phase~\cite{regress2,jiang2013accurate}. Since they do not explicitly represent or use information about the human joint impedance in the controllers, performance deteriorates in practical settings when physical interactions with the environment cause sEMG signals to vary from those observed during training.
Moreover, representing human motor intent only in terms of kinematics makes it difficult to understand the prosthesis' operation. Researchers who seek to achieve simultaneous control of joint position and stiffness typically use muscle-tendon models to decode the human motor intent, and then tune the stiffness extracted from the trained muscle-tendon units (MTUs) in a calibration phase to satisfy the controller's stability constraints~\cite{capsi2020exploring,hocaoglu2022semg,karavas2015tele,niu2021neuromorphic,osu2002short}. This approach creates a mismatch between the MTUs' dynamics and those of the robot, affecting the user's ability to control the prosthesis and limiting the control method's transparency.

We propose a novel framework that includes the user in the control loop and provides three Degrees of Control (DoC), i.e., the control of joint position, stiffness, and damping, of a Degree of Freedom (DoF) of a simulated robot. Our framework comprises two blocks in cascade---see Figure~\ref{fig:framework}-A:
\begin{enumerate}
    \item The first block includes MTUs and maps input sEMG signals to an estimate of the user's motor intent in terms of kinematics (i.e., joint motion) and dynamics (i.e., joint stiffness and damping); 
    
    \item The second block then executes this motor intent through a robot system using a variable position-based impedance controller, allowing online position control and adaptation of the prosthesis' impedance.
\end{enumerate}
The \emph{novelty} is in the support for multiple DoC for each DoF, and in using the stiffness and damping estimated from the MTUs to implement the user's intended behavior on the robot, thus ensuring that the MTUs' dynamics match those of the robot enhancing the framework's transparency.

To fully explore the interplay between the framework's components, we conducted a study with human subjects focusing on the flexion-extension of their wrist to control a single DoF of a simulated robot. We evaluated the framework with eight able-bodied participants and an amputee during online reaching tasks in static environments and in the presence of external perturbations in the form  of a force field pushing the simulated wrist away from the target. We experimentally demonstrate that:  
\begin{itemize}
    \item Our framework supports online adaptation of the simulated robot's kinematics and dynamics in response to external disturbances; the performance is similar to that of the state-of-the-art baseline in the absence of perturbation but significantly better in the presence of perturbations; 
    \item Users' perception of the framework's controllability matches quantitative results; \emph{controllability} here refers to stability and responsiveness to variations in the sEMG signals due to changes in muscle coactivation (i.e. impedance) and to the user's ability to achieve stability after an external perturbation. \emph{There has been limited exploration of the impact of muscle activation modulation on the control of prostheses}.
\end{itemize}
Overall, our study provides interesting insights into the estimation and mapping of user motor intent to control an upper-limb prosthesis, which could be useful in other applications as well.  

\section{Related work}
\label{sec:relatedwork}
It is well-established in the research literature that impedance modulation of muscles is a critical motor-control strategy in humans for stable physical interaction with the environment~\cite{hogan1990mechanical,flash1990human,perreault2001effects}. 
Muscle impedance is a complex concept, but it has been summarised as mechanical impedance, a dynamic operator that defines the forces generated by an object in response to some imposed motion \cite{hogan1985mechanics}. 

Research in robotics has widely used the mechanical impedance paradigm to shape the interaction between a robot and the environment~\cite{hogan1985impedance}. However, implementing variable impedance behaviour on a robot is challenging since the choice of controller gains depends on the task, system constraints, and environmental properties. In human-robot interaction scenarios that require the robot to physically cooperate with humans or acquire the user's control skills (e.g., teleoperation, prosthesis control), it is crucial to allow the user to control the robot's kinematics and dynamics in order to modulate its interaction with the environment. This requirement is more pronounced with prostheses because the user and the prosthesis (i.e., robot) are two agents in the same system.

Surface electromyography (sEMG) is used widely to establish a continuous interface between the user and the prosthesis because it is non-invasive and easy to use. Existing methods that attempt to enable simultaneous control of joint position and impedance of a single DoF, estimate stiffness directly from the magnitude of the EMG signals, even when muscle-tendon models are used within the pipeline and the impedance parameters could be retrieved from the muscle-tendon state~\cite{capsi2020exploring,hocaoglu2022semg,karavas2015tele,niu2021neuromorphic,osu2002short}. In addition, joint damping is often not computed from the sEMG signals or from the muscle-tendon models' state in these methods; it is instead set as a function of the stiffness. This means that the dynamics of the muscle-tendon model does not match that of the robot, possibly leading to unwanted behaviour and instability. Other methods first estimate the stiffness and damping from the measured torque and then learn a model (e.g., a polynomial function) of the stiffness and damping~\cite{furui2019myoelectric,tsuji2010biomimetic}. We argue that these approaches make simplifying assumptions on the relationship between the amplitude of the EMGs and the torque; since they rely on the relationship between EMG signals, the output torque and stiffness are very sensitive to the amplitude and shape of the EMG signals. No commercial upper-limb prosthesis used for daily life tasks allows adaptation of joint impedance properties due to limitations imposed by the hardware design and the challenges in implementing the desired adaptive behaviour. 

Researchers have explored different function approximators for simultaneous and proportional myocontrol of one or more DoFs of upper-limb prostheses~\cite{regress2,regress4,regress5}. Unsupervised methods
have been used to improve robustness to issues such as electrode shifting that induce EMG signal deterioration~\cite{yeung2022co}. Results indicate that non-linear regression outperforms linear regression during the training phase, but results are comparable on unseen testing data~\cite{krasoulis2015evaluation}. In recent years, deep learning methods are increasingly being used to improve feature extraction and learn complex non-linear mappings between EMGs and target motion trajectories or classes~\cite{ann1,vujaklija2018online}. While these data-driven methods can support simultaneous and proportional control of two DoFs, they do not explicitly model the human motor intent in terms of joint stiffness and damping, and do not allow the user to adapt the prosthesis' impedance in the presence of new perturbations (e.g., external force fields). Moreover, there is lack of evaluation of these methods in scenarios where impedance adaptation is required to maintain system stability.

\section{Methodology}
\label{sec:framework}
This section describes our framework and its components in Sections~\ref{sec:framework-overview}-~\ref{sec:framework-prosthesis-control}), and the optimization method used to train the muscle-tendon models (Section~\ref{sec:framework-learn-params}). 
In the description below, the time dependence of variables is dropped for simplicity.

\subsection{Overview of framework architecture}
\label{sec:framework-overview}
Figure~\ref{fig:framework}-A provides an overview of our framework in the context of controlling a single DoF through wrist flexion and extension.

The \textbf{\textit{Detection of Human Motor Intent}} block takes as input the sEMG signals from the user's wrist flexor and extensor muscles ($ch_{1}, ch_{2}$) and estimates the user's motor intent in terms of reference kinematics $s_{r} = (q_{r},\dot{q}_{r},\ddot{q}_{r})$, and joint stiffness $K$ and damping $D$. This estimation is done in two phases. In \textbf{\textit{phase 1}}, the sEMG signals drive the lumped muscle-tendon units ($MTU_{1}$, $MTU_{2}$) that generate the muscle-tendon forces $(F_{1}, F_{2})$ based on the muscle-tendon contraction dynamics and $s_{r}$. Note that the MTUs are virtually arranged on the link of the simulated robot model; their state and ${q}_{r}$ are linked to each other (Figure~\ref{fig:framework}-B). The muscle-tendon stiffness and damping are computed from the MTUs state. These MTUs variables are then mapped to the robot model's joint space to obtain the joint torque $\tau_{r}$, stiffness $K$, and damping $D$. In \textbf{\textit{phase 2}}, $\tau_{r}$ is applied to the simulated robot model to actuate the robot's joint $q_{r}$ and obtain the user intended joint kinematics $s_{r}$. 

The \textbf{\textit{Prosthesis control}} block executes the motor intent obtained from the previous block on the robot plant using a position-based variable impedance controller that tracks $s_{r}$ based on $K$ and $D$. The joint position $q_{f}$ is the framework's output and visual feedback. If an external perturbation acts on the robot plant, the user can use this feedback to modulate the plant's kinematics and impedance, reducing the error between $s_{r}$ and the plant's state. 

In this paper, sEMG signals are processed using proprietary software, and the framework is implemented using CoppeliaSim simulation environment~\cite{6696520} and MATLAB~\cite{MATLAB:R2018b_u8}. We use the model of the\textit{Puma 560} robot as the robot arm because its characteristics are well-understood; we only consider the chain from link 0 to link 2 of this robot and control joint 2 with our framework. In practice, this framework can be used to control a single DoF of any robot. In this work, we do not use a real robot (i.e., prosthesis), so the robot plant is simulated. When the framework is used with a real prosthesis, the \textit{simulated robot model} in Figure~\ref{fig:framework}-A will be a simulation of the prosthesis and the \emph{robot plant} will be the prosthesis.

\subsection{Detection of human motor intent}
\label{sec:framework-motor-intent}
The first block of our framework maps the input sEMG signals ($ch_{1}, ch_{2}$) to an estimate of the user's motion intent ($s_{r}$, $K$, $D$). 

\subsubsection{Muscle-tendon model basics}
The biomechanical model of each $MTU_{i}$ is based on the well-established Hill's muscle-tendon model~\cite{hill1938heat}. Specifically, we adopt the model structure discussed in~\cite{gunther2007high} which showed how the serial damping element DE enables the suppression of high-frequency oscillations within the model. Each $MTU_{i}$ of length $l_{MTU}^{i}$ is composed of a muscle of length $l_{ce}^{i}$ in series to a tendon of length $l_{se}^{i}$. The muscle is modeled by a contractile element (CE) and a parallel elastic element (PE). The tendon is made up of a serial elastic element (SE) in parallel to a damper element (DE). The system at equilibrium is described by:
\begin{equation}
\label{eq:equil}
\begin{aligned}[b]
  	 F_{ce}^{i}(l_{ce}^{i},\dot{l}_{ce}^{i}, ch_{i}) + F_{pe}(l_{ce}^{i}) = F_{se}^{i}(l_{ce}^{i},l_{MTC}^{i}) + F_{de}^{i}(l_{ce}^{i},\dot{l}_{ce}^{i},\dot{l}_{MTC}^{i}, ch_{i})
\end{aligned}
\end{equation}
where the EMG signal $ch_{i}$ is normalised in the range $(0, 1]$ with respect to the maximum value generated by the subject during the training phase.
For detailed information about the numerical simulation of the muscle-tendon contraction dynamics and activation dynamics please see~\cite{buchanan2004neuromusculoskeletal,gunther2007high}.

\subsubsection{Geometric arrangement of MTUs on robot's link}
Figure~\ref{fig:framework}-B provides an overview of the arrangement of the MTUs on the simulated robot model in our framework. Each $MTU_{i}$ is virtually attached to the link from the Center of Mass (CoM) of the link ($l_a^{i}$) to a fixed base ($l_b^{i}$), i.e., the length $l_{MTU}^{i}$ is a function of $q_{r}$. Given $\alpha_{i}$ and the initial joint position $q_{r} = 0$,  we can compute $l_{a}^{i}$ as $l_{MTU}^{i}\sin{\alpha_{i}}$. Also, the values of $l_{a}^{i}$ and $l_{b}^{i}$ are constant during simulation and control, while $l_{MTU}^{i}(q_{r})$ and moment arm $r^{i}(q_{r})$ are a function of $q_{r}$:
\begin{equation}
    l_{MTU}^{i} = \sqrt{(l_{a}^{i})^{2} + (l_{b}^{i})^{2} - 2l_{a}^{i}l_{b}^{i}\cos{(\pi/2 - q_{r})}}
\end{equation}
where the muscle-tendon contraction velocity $\dot{l}_{MTU}^{i}$ is computed by numerical differentiation. Next, to transform the muscle-tendon forces, stiffness, and damping to the joint space, we define the Jacobian matrix \textbf{R} = $[r^{1}  \ r^{2}]$ =
$[\pdv{l_{MTU}^{1}(q_{r})}{q_{r}} \,\,\, \pdv{l_{MTU}^{2}(q_{r})}{q_{r}}]^T$ containing the moment arms $r^{i}$ of the two MTUs:
\begin{equation}
\begin{aligned}
    & r^{i}(q_{r}) = \pdv{l_{MTU}^{i}(q_{r})}{q_{r}} = l_{b}^{i}\sin{\alpha^{i}(q_{r})} 
    \\& \textrm{with} \quad 
    \alpha^{i}(q_{r}) = \acos{(\frac{-(l_{a}^{i})^{2} + (l_{b}^{i})^{2} + (l_{MTU}^{i})^{2}}{2l_{MTU}^{i}l_{b}^{i}})}
\end{aligned}
\end{equation}

\subsubsection{Joint stiffness and damping estimation}
The stiffness and damping of each MTU are first estimated from the MTUs state and then mapped to the simulated robot's joint space. The stiffness $K_{i}$ is modeled as the muscle fiber stiffness $K_{m}^{i}$ in series with the tendon's stiffness $K_{t}^{i}$, i.e., $K_{i} = K_{m}^{i}K_{t}^{i}/({K_{m}^{i} + K_{t}^{i}})$. Among the approaches available to compute the muscle stiffness (see Section~\ref{sec:relatedwork}), we chose the methods that computes $K_{m}^{i}$ as the directional derivative of $F_{ce}(l_{ce}^{i},\dot{l}_{ce}^{i},ch_{i})$ with respect to unit vector of $l_{ce}^{i}$~\cite{sartori2012emg}:
\begin{equation}
 K_{m}^{i} = \pdv{F_{m}^{i}(l_{ce}^{i},\dot{l}_{CE}^{i},ch_{i})}{l_{ce}^{i}}
\end{equation}
This formulation takes into account the state of the muscle ($l_{ce}^{i},\dot{l}_{ce}^{i},ch_{i}$); in the stiffness computation, it removes the force component due to changes in $\dot{l}_{ce}^{i}$ and $ch_{i}$. Similarly, $K_{t}^{i}$ is computed as the directional derivative of $F_{t}^{i} = F_{se}^{i} + F_{de}^{i}$ with respect to unit vector of $l_{se}^{i} = l_{MTC}^{i} - l_{ce}^{i}$.  Muscle and tendon damping ($D_{m}^{i}$, $D_{t}^{i}$) are computed as directional derivatives with respect to muscle contraction velocity $\dot{l}_{ce}^{i}$ unit vector and tendon extension velocity $\dot{l}_{se}^{i} =\dot{l}_{MTU}^{i} - \dot{l}_{ce}^{i}$.

\subsubsection{Mapping from muscle space to joint space}
Forces generated by the MTUs result in the net torque $\tau_{r} = [F_{1}, F_{2}]^{T}\textbf{R}$.
We compute the joint space stiffness considering also the contribution of the moment arms as a function of $q_{r}$~\cite{hogan1990mechanical}:
\begin{equation}
        K = \pdv{\tau_{r}}{q_{r}} = \pdv{\textbf{R}^{T}}{q}[F_{1},F_{2}]^{T} + \textbf{R}^{T}diag([K_{1},K_{2}])\textbf{R}
\end{equation}
The joint damping is computed as $D = \sum_{i=1}^2(D_{i}(r^{i})^{2})$.

\subsection{Prosthesis control}
\label{sec:framework-prosthesis-control}
At this point, human motor intent at time $t$ is represented by the joint torque $\tau_{r}(t)$, stiffness $K(t)$, and damping $D(t)$. To implement this intent using the impedance control, the intended joint kinematics has to be retrieved. We use the forward dynamics of the simulated robot model to obtain the joint state $s_{r}(t) = ({q}_{r}(t),\dot{q}_{r}(t),\ddot{q}_{r}(t)$) from \textbf{$\tau_{r}(t)$}. A position-based variable impedance controller is then used to track $s_{r}(t)$ by adapting $K(t)$ and $D(t)$. The dynamics model for a robot with one rotational joint is:
\begin{equation}
    M\ddot{q}_{r}(t) + g(q_{r}(t)) = \tau_{f}(t) + \tau_{ext}(t)
\end{equation}
where M is the link's joint space inertia, $g$ is the gravity compensation torque, and $\tau_{ext}(t)$ is the external perturbation on the robot joint. We build on the impedance control method used in the absence of force-torque readings~\cite{hogan1985impedance} to define the control law:
\begin{equation}
	\tau_{f} = M\ddot{q}_{r}(t) + K(q_{r}(t) - q_{f}(t)) + D(\dot{q}_{r}(t) - \dot{q_{f}}(t)) + g(q_{r}(t))
\end{equation}
Note that this definition uses the robot's link inertia since only low accelerations are reached during control. 
We design the MTUs' length and contraction velocity to be a function of $s_{r}$ so that any external perturbation ($\tau_{ext}$) only affects $s_{f}$; the state of the simulated robot model $s_{r}$ and that of the MTUs are unaffected and represent the user motor intent based on the input sEMG signal.
This choice enables the implementation of the user's "corrective" action as the feedback loop of the impedance controller. In the absence of external perturbations ($\tau_{ext} = 0$), $q_{r}$ matches $q_{f}$. If $\tau_{ext}$ is non-zero, depending on $K$ and $D$, $q_{f}$ will start diverging from $q_{r}$. This $q_{f}$ serves as visual feedback for the user, who can perform run-time adaptation of the simulated robot's state and gains (K, D) by modulating the muscles' coactivation.

\subsection{Muscle-tendon models training}
\label{sec:framework-learn-params}
For our framework to work as described above, it is important to set suitable values for the parameters of the MTUs. Table~\ref{tab:params} lists the parameters $\overline{\boldsymbol{p}}^{i} \in \mathbb{R}^{m}$ of $MTU_{i}$ to be optimized. Typically, the underlying optimization process minimizes the root mean square error (RMSE) between a reference torque and $\tau_{r}$ without any prior knowledge of the interaction between elements of the MTUs; this makes the optimization of the MTU parameters an ill-posed problem. Multiple combinations of parameters can generate the same net joint torque $\tau_{ref}$ and thus the same reference joint position $q_{r}$, but most combinations may provide gains (K, D) that do not match the user-desired dynamic behaviour and can result in an ineffective impedance controller. This is a critical issue that undermines transparency, making it difficult to understand the MTUs' dynamics and to extract valid control gains values. Toward addressing this issue, we:

\begin{table}[tb]
\caption{Parameters $\overline{\boldsymbol{p}}^{i}$ defining $MTU^{i}$; see~\cite{gunther2007high} for details. The column "variable" lists the m parameters as a function of the variables $\overline{\boldsymbol{p}}^{i}$ to be optimized. The lower and upper bound of each parameter is indicated in the last two columns and set experimentally and based on prior work~\cite{scovil2006sensitivity}.}
\label{tab:params}
\resizebox{\columnwidth}{!}{%
\begin{tabular}{lllll|l|l|l|l|}
\cline{1-4} \cline{6-9}
\multicolumn{1}{|l|}{\textbf{\begin{tabular}[c]{@{}l@{}}Parameter \\ Name\end{tabular}}} &
  \multicolumn{1}{l|}{\textbf{Variable}} &
  \multicolumn{1}{l|}{\textbf{\begin{tabular}[c]{@{}l@{}}Lower \\ Bound\end{tabular}}} &
  \multicolumn{1}{l|}{\textbf{\begin{tabular}[c]{@{}l@{}}Upper \\ Bound\end{tabular}}} &
   &
  \textbf{\begin{tabular}[c]{@{}l@{}}Parameter \\ Name\end{tabular}} &
  \textbf{Variable} &
  \textbf{\begin{tabular}[c]{@{}l@{}}Lower \\ Bound\end{tabular}} &
  \textbf{\begin{tabular}[c]{@{}l@{}}Upper \\ Bound\end{tabular}} \\ \cline{1-4} \cline{6-9} 
\multicolumn{1}{|l|}{$F_{max}$} &
  \multicolumn{1}{l|}{$\bar{p}_{1}$} &
  \multicolumn{1}{l|}{1000} &
  \multicolumn{1}{l|}{9000} &
   &
  $v_{pee}$ &
  $\bar{p}_{10}$ &
  1.1 &
  3 \\ \cline{1-4} \cline{6-9} 
\multicolumn{1}{|l|}{$l_{opt}$} &
  \multicolumn{1}{l|}{$\bar{p}_{2}l_{ce}^{init}$} &
  \multicolumn{1}{l|}{0.05$l_{ce}^{init}$} &
  \multicolumn{1}{l|}{0.085$l_{ce}^{init}$} &
   &
  $f_{pee0}$ &
  $\bar{p}_{1}\bar{p}_{11}$ &
  0.5$\bar{p}_{1}$ &
  1$\bar{p}_{1}$ \\ \cline{1-4} \cline{6-9} 
\multicolumn{1}{|l|}{$W_{des}$} &
  \multicolumn{1}{l|}{$\bar{p}_{2}\bar{p}_{3}$} &
  \multicolumn{1}{l|}{0.7$\bar{p}_{2}$} &
  \multicolumn{1}{l|}{3.5$\bar{p}_{2}$} &
   &
  D &
  $\bar{p}_{12}$ &
  0.001 &
  3 \\ \cline{1-4} \cline{6-9} 
\multicolumn{1}{|l|}{$W_{asc}$} &
  \multicolumn{1}{l|}{$\bar{p}_{2}\bar{p}_{4}$} &
  \multicolumn{1}{l|}{0.7$\bar{p}_{2}$} &
  \multicolumn{1}{l|}{3.5$\bar{p}_{2}$} &
   &
  R &
  $\bar{p}_{13}$ &
  0 &
  0.8 \\ \cline{1-4} \cline{6-9} 
\multicolumn{1}{|l|}{$v_{des}$} &
  \multicolumn{1}{l|}{$\bar{p}_{5}$} &
  \multicolumn{1}{l|}{1.2} &
  \multicolumn{1}{l|}{3} &
   &
  $l_{see0}$ &
  $\frac{2}{3}l_{MTU}$ &
  $\frac{2}{3}l_{MTU}$ &
  $\frac{2}{3}l_{MTU}$ \\ \cline{1-4} \cline{6-9} 
\multicolumn{1}{|l|}{$v_{asc}$} &
  \multicolumn{1}{l|}{$\bar{p}_{6}$} &
  \multicolumn{1}{l|}{1.2} &
  \multicolumn{1}{l|}{3} &
   &
  $\Delta Unl$ &
  $\bar{p}_{14}$ &
  0.02 &
  0.07 \\ \cline{1-4} \cline{6-9} 
\multicolumn{1}{|l|}{$A_{max}$} &
  \multicolumn{1}{l|}{$\bar{p}_{7}$} &
  \multicolumn{1}{l|}{0.1} &
  \multicolumn{1}{l|}{0.4} &
   &
  $\Delta Ul$ &
  $\bar{p}_{14}\bar{p}_{15}$ &
  $\frac{1}{3}\bar{p}_{15}$ &
  $\frac{2}{3}\bar{p}_{15}$ \\ \cline{1-4} \cline{6-9} 
\multicolumn{1}{|l|}{$B_{max}$} &
  \multicolumn{1}{l|}{$\bar{p}_{8}$} &
  \multicolumn{1}{l|}{1.1} &
  \multicolumn{1}{l|}{5.1} &
   &
  $\Delta F_{see0}$ &
  $\bar{p}_{1}\bar{p}_{16}$ &
  0.3$\bar{p}_{1}$ &
  1$\bar{p}_{1}$ \\ \cline{1-4} \cline{6-9} 
\multicolumn{1}{|l|}{$l_{pee0}$} &
  \multicolumn{1}{l|}{$\bar{p}_{2}\bar{p}_{9}$} &
  \multicolumn{1}{l|}{0.7$\bar{p}_{2}$} &
  \multicolumn{1}{l|}{0.95$\bar{p}_{2}$} &
   &
  S &
  $\bar{p}_{17}$ &
  1.2 &
  2 \\ \cline{1-4} \cline{6-9} 
 &
   &
   &
   &
   &
  F &
  $\bar{p}_{18}$ &
  0.5 &
  2 \\ \cline{6-9} 
\end{tabular}%
}
\vspace{-1.5em}
\end{table}

\begin{enumerate}
\item make assumptions about the structure of MTUs;
\item reduce the number of parameters based on relative importance and maximize the parameters' interdependence; 
\item define an optimization method that includes an impedance controller and uses $q_{f}$ as the optimization signal.
\end{enumerate}
\paragraph{(1) Structural assumptions on MTU}
The MTU's parameters influence the interaction between the elements in the muscle and in the tendon, which (in turn) affects their stiffness and damping. We found in initial sensitivity studies that the result of the optimization is highly sensitive to the relative lengths of the muscle and the tendon since it affects how each element of the MTU contributes to the overall force generated by the MTU. Existing literature investigating the relationship between architectural features of the MTUs and their contraction dynamics indicates that muscle-tendon systems characterised by a longer tendon than the muscle enhance control and impedance modulation~\cite{winters1990hill,bennett11986mechanical}. This hypothesis about the functional properties enabled by this MTU's structure has been investigated in~\cite{rack1984tendon}.
Based on these studies, our framework's MTUs have a long tendon compared to the muscle, with the tendon slack length of $\frac{2}{3}l_{MTU}^{i}$. The ratio between the muscle and tendon length matches that of the muscle-tendon complex investigated in~\cite{rack1984tendon}.

\paragraph{(2) Simplify MTU's parameters}
Sensitivity studies indicated that some parameters did not make any substantial contribution to the offline tracking ability. We made two simplifications to the MTU parameters to be estimated: (i) the pennation angle, i.e., the angle between the muscle and the tendon, is set to be constant at zero; and (ii) 
$l_{opt}^i$ is learned directly from data and assumed to be constant instead of being a function of the input activation~\cite{lloyd2003emg,karavas2015tele}.

\paragraph{(3) Optimization signal} Instead of the torque $\tau_{r}$, we used the final joint position $q_{f}$, which depends on the dynamics defined by the gains ($K$, $D$), as the optimisation signal. \emph{This important change enabled us to train the MTUs such that the stiffness and damping estimated from the MTUs' state can be incorporated directly into the position-based variable impedance controller without further tuning}.
The prediction function $f:\mathbb{R}^{2m+2}\rightarrow \mathbb{R}$ acts on the input defined by the sEMG signals $[ch_{1}(t),ch_{2}(t)] \in \mathbb{R}^{2}$ and the parameters of the MTUs $\overline{\textbf{p}}=[\overline{\textbf{p}}^{1},\overline{\textbf{p}}^{2}] \in \mathbb{R}^{2m}$ to produce the final joint position ${q}_{f}(t)\in \mathbb{R}$.
Then, the constrained optimization problem is: 
\begin{equation}
\begin{aligned}
\min_{\overline{\textbf{p}}} \quad & \sqrt{\frac{\sum_{t=1}^{T}(f([ch_{1}(t),ch_{2}(t)];\overline{\textbf{p}})- q_{f}^{train}(t))^2}{T}} \\
\textrm{s.t.} \quad & \textbf{lb} \leq \overline{\textbf{p}} \leq \textbf{ub}\\
\end{aligned}
\end{equation}
where $q_{f}^{train}(t)\in \mathbb{R}$ is the measured wrist flexion-extension angular position; $\textbf{lb},\textbf{ub}\in \mathbb{R}^{2m}$ are the lower and upper bounds of $\overline{\textbf{p}}$ in Table~\ref{tab:params}, and $T$ is the length of the trajectories. Additionally, the following constraints are added to the optimization problem to prevent numerical instability and aid in convergence:
\begin{itemize}
    \item $W_{des}^{i}+W_{asc}^{i} < l_{ceInit}^{i}$, where $l_{ceInit}^{i}$ is $l_{ce}^{i}$ when $q_{r} = 0$, such that CE operates in the muscle-length range.
    \item if $l_{ce}^{i} < 0.001l_{opt}^{i}$ or $l_{ce}^{i} > 0.95(l_{MTU}^{i} - l_{see0}^{i})$ set $\dot{l}_{ce}^{i} = 0$ such that $l_{MTU}^{i} = l_{ce}^{i} + l_{see0}^{i}$ and tendon cannot be compressed.
    \item $K > 0, D\geq0$; required for control stability.
    \item the maximum extension of the tendon ($l_{se}^{i}$) is $0.1\cdot l_{see0}^{i}$~\cite{winters1990hill}.
\end{itemize}

\section{Experimental evaluation}
\label{sec:result}
The experimental setup and experimental protocols are illustrated in Figure~\ref{fig:exp} and described below. We also describe the baseline, performance measures, the human participants group, and the hypotheses being evaluated. 

\subsection{Experimental setup}
\label{sec:result-setup}
\begin{figure*}
\includegraphics[width=1\textwidth]{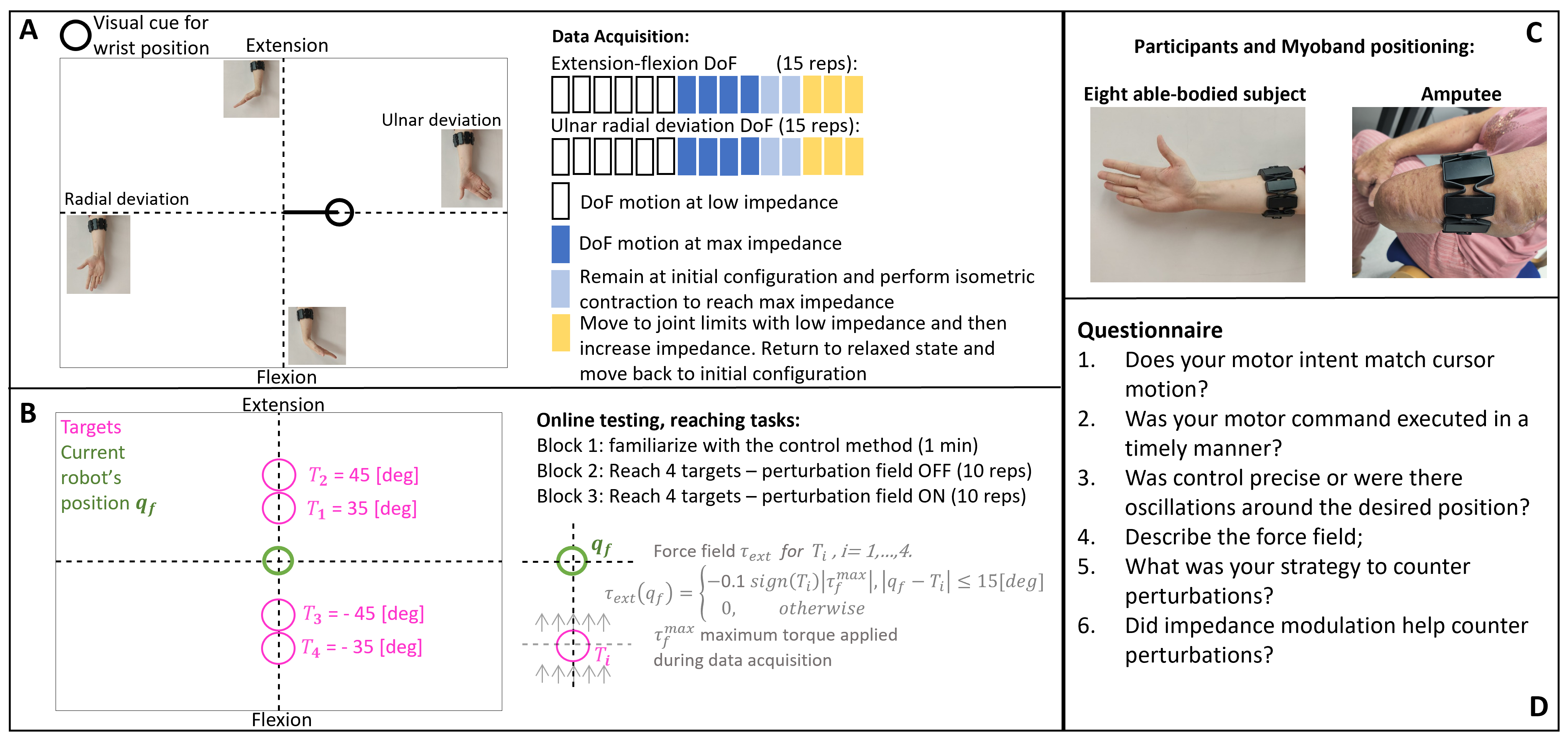}
\vspace{-1em}
\caption{\textbf{A)} Data acquisition: subject's wrist motions are guided by a visual cue (black circle), and the data acquisition protocol. \textbf{B)} Reaching task: subject receives visual feedback about the robot plant's joint position $q_{f}$ (green circle) and has to reach targets T1-T4 using the experimental protocol (on right). \textbf{C)} Position of EMG sensors on the subject's forearm. D) Subject questionnaire.}
\label{fig:exp}
\end{figure*}

During the experiments, each human subject sat in front of a screen, with their arm along the side of the body in a neutral resting position. They wore a Myoband by ThalmicLab (eight sEMG channels, frequency 200Hz) to record the muscle activations (\textbf{E}(t)) as they tried to reach the target (visual cue) shown on screen---Figure~\ref{fig:exp}-A. Our long-term target population is transradial amputees (i.e. loss of limb below the elbow); the myoband was thus positioned $\approx 5$ cm below the elbow for all participants (Figure~\ref{fig:exp}-C). We considered channels 4 and 8 as the EMG signals of the main extensor and flexor muscle ($ch_{1}, ch_{2}$). The wrist position $q_{f}^{train}$ was tracked with a Qualysis motion capture system for the able-bodied subject. For the amputee, $q_{f}^{train}$ is the trajectory of the visual cue the participant has to follow during the training experiment. The raw EMG signals were bandpass-filtered (20 - 500 Hz) and full-wave rectified before the root-mean-square temporal features were extracted (moving window of length 160ms and step size 40 ms). All experiments were run on a computer with an Intel i7-8809G, 3.10GHz CPU and a Radeon RX Vega M GH graphic card.

\subsection{Experimental protocol}
For each participant, experiments were conducted in three sessions. In the first session (offline training), we collected data to train our framework M and baseline B; M and B are then tested on separate days (online reaching task) to avoid muscle fatigue and involuntary bias due to the order in which frameworks were evaluated. 

\subsubsection{Data acquisition and model training}
As shown in Figure~\ref{fig:exp}-A, the subject viewed a screen with Cartesian space axes corresponding to wrist joint positions (in \textit{Degrees}) for ulnar-radial deviation and flexion-extension respectively. During each trial, a visual cue moved along one of the axes and the subject had to move their wrist to proportionally match this cue. Each DoF motion is repeated 15 times, while the subject is instructed to perform the wrist motion while modulating the impedance properties of the muscles---see experimental protocol in Figure~\ref{fig:exp}-A.
Even when we are primarily activating a single DoF (e.g.,  ulnar-radial deviation), there is some unintentional motion along the other DoF (i.e., flexion-extension). So, although we focus on the control of a single DoF (flexion-extension), we asked the subject (during data collection) to also perform repetitions of ulnar-radial deviation so that we could observe the indirect flexion-extension motion. Data (i.e., EMG signals, wrist position $\textbf{q}_{f}^{train}$) from $15$ trials of each kind of motion (i.e., 30 trials total) were collected. A 60-40 split of this data was used for training and validating the muscle models, with optimization based on Simulated Annealing (500 iterations, 5000 function evaluations, initial value of temperature 300, annealing interval 50)~\cite{van1987simulated} since the cost function has discontinuous derivatives. The same overall process is followed for able-bodied participants and an amputee, except that the amputee is not asked to perform ulnar-radial deviations given how difficult it is for them to learn to generate more than two independent muscle activations that can be used as control signals.

\subsubsection{Online testing: reaching task}
In each trial of online testing, a participant had visual feedback of their predicted wrist position (i.e., $q_{f}$, green circle) and was asked to perform wrist flexion-extension to reach a target position $T_{i}$ (purple circle) as accurately and quickly as possible (Figure~\ref{fig:exp}-B). Once at the target, the subject had to maintain the position for three seconds. Every time the subject enters the target circle but can't maintain the position (the wrist position circle intersects the target circle) the 3-second dwelling time is reset. The radius of the circle indicating $q_{f}$ and the target $T_{i}$, were eight and six density-independent pixels respectively, requiring precise control. Experimental trials for each subject were divided into three blocks (Figure~\ref{fig:exp}-B): familiarisation with the control interface; reaching tasks in the free space; and reaching tasks in the presence of a perturbation field $\tau_{ext}$ that pushed $q_{f}$ away from the target.

At the beginning of each session, the subject was told that different motor control strategies could be explored, e.g., relaxed movement and changing muscle co-activation, but the subject had no prior knowledge of the method (M or B) being tested. The subject was told that some force would perturb $q_{f}$, but no information about the force field (type, magnitude, location) was provided. The magnitude of the perturbation was defined as a percentage of the maximum torque $\tau_{f}$ generated by the subject during training; its average, across all participants, was $\approx 20 Nm$.

\subsection{Data-driven baseline}
The baseline framework built on a neural network (NN) that learned a mapping from EMG signals to the reference kinematics~\cite{jiang2013accurate}. The same training data was used for our framework and the baseline that was trained to match the performance reported in~\cite{jiang2013accurate}. To ensure accurate motion tracking and perturbation rejection, a high-stiffness PD controller was added in cascade to the NN to track the joint position $q_{r}$ predicted by the NN. This controller had constant high stiffness ($K_{B} = 100 [N/rad]$) and damping computed assuming a critically damped system ($D_{B} = \sqrt{K_{B}/4}  [Ns/rad]$)~\cite{kronander2013learning}. Also, $q_{f}$ was the output position when tracking $q_{r}$ with $K_{B}$ and $D_{B}$. The choice of baseline was motivated by the fact that NN-based (regression) methods are the state of the art for the control of prostheses. Also, existing methods that do include MTUs do not directly use the stiffness computed by the MTUs; the joint stiffness is instead tuned for stable position tracking during a separate calibration phase, which is, in principle, equivalent to the use of a high-stiffness position controller in our baseline.  

\subsection{Performance measures}
Next, we describe the performance measures used to evaluate the outcomes provided by M and B, 
the questionnaire used to survey participants' perceived controllability of each framework with and without any external perturbations, and the composition of the human subjects involved in this study.  

\subsubsection{Performance measures}
We selected six performance measures widely used to evaluate prostheses control methods~\cite{williams2008evaluation}: (i) \textit{Success Rate} (\textbf{SR}) [\%]: proportion of successful trials, with a trial considered successful if the subject reached the target within 30 seconds and held the position for three seconds; (ii) \textit{Time to Reach} (\textbf{TR}) [sec]: time to complete the trial, with 30 seconds as the maximum allowed time; (iii) \textit{Throughput} (\textbf{TP}) [bit/sec]: $\frac{ID}{TR}$ quantifies the information conveyed during TR, considering distance to target and the target's radius; (iv) \textit{Path Efficiency} (\textbf{PE}) [\%]: the ratio between the length of optimal path to target (Euclidean distance) and the length of the trajectory executed by the subject; (v) \textit{Energy} [J]: $\int_{t_i}^{t_f}\tau_{f}(t)\dot{q}_{f}\;\mathrm{d}t$, where ($t_f - t_i$) is the trial duration; and (vi) \textit{Near Miss} (\textbf{NM}) [\#]: number of times the subject entered the target circle, but did not maintain the position for three seconds.

We computed the \textit{smoothness} of $q_{f}$ using the \textbf{SPARC} measure~\cite{balasubramanian2015analysis}; a higher value is obtained if the subject successfully counters the external perturbations. We also computed the \textit{Mutual Information} (\textbf{MI}) between $\tau_{f}$ and $q_{r}$ to quantify the predictability of $q_{r}$ given $\tau_{f}$; MI has been used in literature for dynamic system analysis (e.g.~\cite{bazzi2020human}). Since $q_r$ is the unperturbed reference trajectory and $\tau_f$ is the torque that results in $q_f$, we expect MI to increase when $q_r$ matches $q_f$, i.e., when the subject quickly counters the perturbation. We also computed the conditional MI between $q_{r}$ and $q_{f}$ given K and D respectively, to provide insights into the contribution of stiffness and damping to the reduction in position error $q_{r}-q_{f}$ of the impedance controller. 
The one-tailed Mann–Whitney U test was used to measure the statistical significance (p-values $<$ 0.05) of the difference in performance provided by M and B.

\subsubsection{Survey of user's perception of controllability}
We explored the user's \textit{perceived controllability} provided by M and B. Subjects were asked to answer six questions about M and B at the end of each experimental session; please see the \emph{Questionnaire} in Figure~\ref{fig:exp}-D.
Users had to choose from one of three options (mostly good, good with some instances of low controllability, and poor) for the first three questions; the other questions allowed free-form answers.

\subsection{Human participants} 
Eight able-bodied, right-handed volunteers (five females, three males, age: $27.87 \pm 3.64$) without neuromuscular disorders and prior experience in myocontrol, took part in the study approved by our university's ethics committee. One upper-limb amputee (female, age 65), who does not use a prosthesis and whose physical condition makes it harder for her to perceive differences in stiffness, took part in the experiments. We evaluated the able-bodied subjects' and the amputee's performance separately, and provide a comparative discussion. For each subject, we trained M and B using the data acquired in the training phase. For the amputee, an engineered reference trajectory of the wrist flexion-extension was used to train M and B. All subjects completed the online reaching task experiment with M and B and then the questionnaire.  The training and evaluation took about two hours, spread over two days, for each able-bodied person; $\approx 30$ minutes for initial preparation and data collection, and $\approx 90$ minutes for online testing with our framework and the baseline. This process took $\approx 6$hours for the amputee because multiple mental and physical factors make it harder for them to learn to express intent. The time commitment requirement, the availability of amputees, and the COVID-19 pandemic strongly influenced the size and composition of the study group.

\subsection{Hypotheses}
We experimentally evaluated the following hypotheses:
\begin{itemize}
    \vspace{-0.5em}
    \item[\textbf{(H1)}] M achieves comparable or higher performance than B in the absence of a perturbation field, and outperforms B when the user has to interact with perturbations;
    
    \item[\textbf{(H2)}] Regulation of joint stiffness and damping is an effective strategy to adapt to perturbations; and

        
    \item[\textbf{(H3)}] The user's perceived controllability is higher with M than with B, especially when the user has to interact with external perturbations.
\end{itemize} 

\section{EXPERIMENTAL RESULTS}
\label{sec:result-analysis}
In this section we discuss key results of the experimental evaluation; additional results, e.g., about offline tracking and time evolution of relevant MTUs, are discussed in the \textit{Supplementary Material}. 
\subsection{Results for able-bodied subjects}
\begin{figure*}
\centering
\includegraphics[width=1\textwidth]{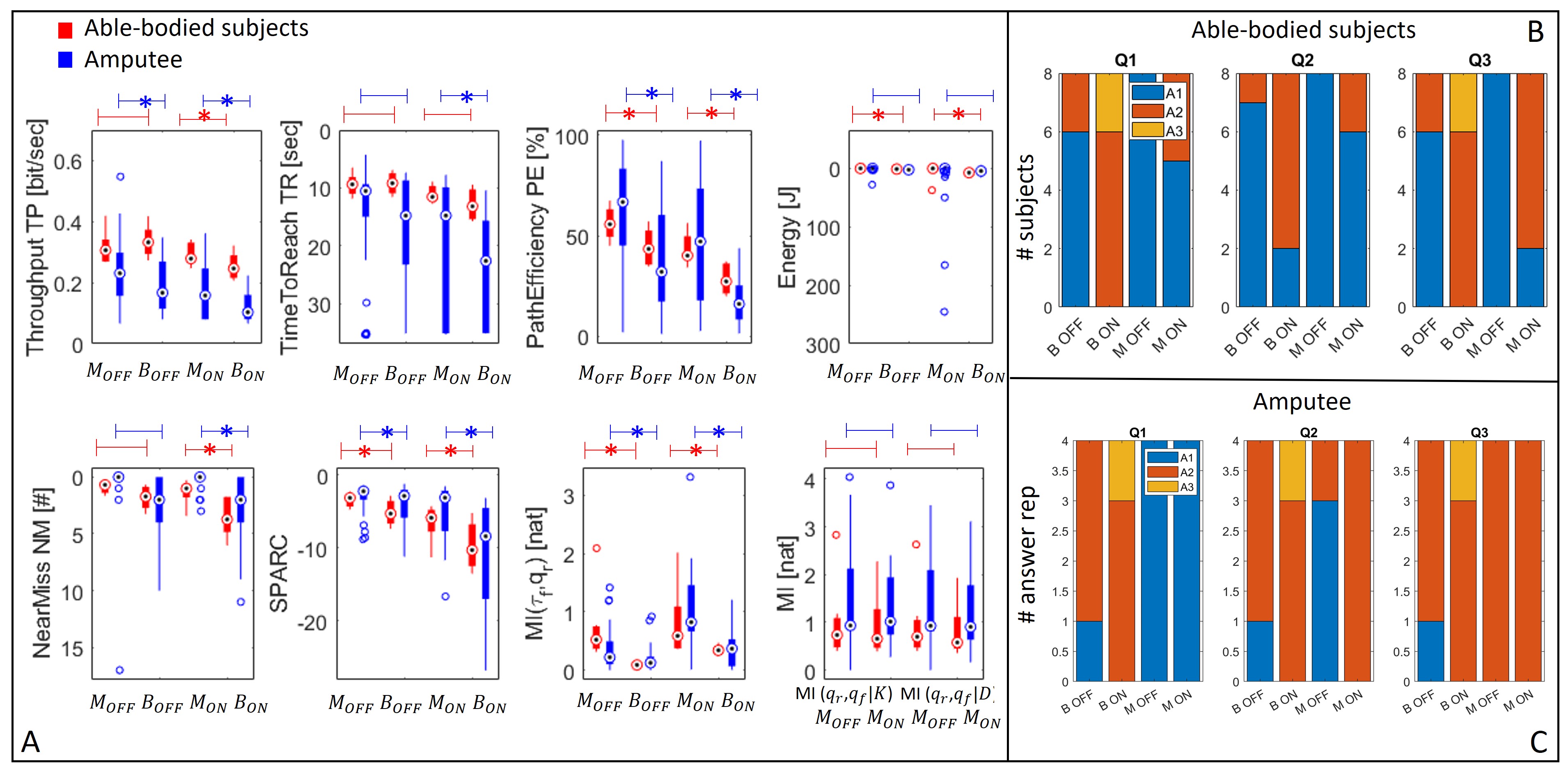}
\vspace{-1em}
\caption{A) Values of performance measures for the able-bodied subjects (red) and the amputee (blue): each red group contains the average performance of the eight subjects (averaged over 40 trials); each blue group contains the 40 trials for the amputee. A statistically significant improvement of the median, computed between M OFF-B OFF and M ON-B ON, is highlighted with  an asterisk. B) Able-bodied volunteers' responses to Q1-3 of the questionnaire, completed at the end of the session with or without external perturbations. Each category shows the fraction of subjects who provided a certain answer. C) Amputee's responses to Q1-Q3; the participant was requested to answer questions after every 10 trials, i.e., four times per session.}
\label{fig:metrics}
\end{figure*}

Figure~\ref{fig:metrics}-A includes the results of evaluating \textbf{H1}. The performance of each of the eight able-bodied participants is shown in red, averaged over the 40 trials  for that participant because there may be bias/dependence between individual trials of a participant. Also, instances of M providing a statistically significant improvement over B are shown with a red asterisk at the top of the plot for the corresponding measure. Unlike B, M consistently enabled successful task completion with or without perturbation; the SR metric was $95\%$ and $82.19\%$ for M and B (respectively) without perturbation, and $93.75\%$ and $76.87\%$ with perturbation. While M and B provided comparable values for TR and TP in the absence of perturbations, M provides a significant improvement in TP with perturbations. Also, M has a significantly higher PE than B with and without perturbations; this matches the much lower NM of M than B (with or without perturbations). Even the energy used by M is significantly lower than that by B, with or without perturbations, which matches the lower TP, longer TR, and lower PE of B than M. These results support \textbf{H1}, i.e., that M outperforms B for reaching tasks in free space and in the presence of external perturbations.

Next, we evaluated whether online modulation of joint stiffness and damping allowed participants to counter perturbations (\textbf{H2}). The value of SPARC was significantly better with M than with B, with or without perturbations, i.e., participants were better able to smooth out the oscillations imposed by perturbations when using M. The smoothness decreased (as expected) for both M and B when perturbations were included.
One explanation for the variance in SPARC when using M (with perturbations) is that the perturbation initially had an impact if the subject was operating with low impedance, but the subject then adjusted the gains suitably. When using B, on the other hand, gains remained at a fixed high value, which enabled rejection of perturbations and accurate tracking but the impedance controller also attempted to track inaccurate estimates of the intended joint position provided by the baseline (see Section 3.1 of Supplementary Material). We observed that M provided a significantly higher MI between $\tau_f$ and $q_r$ compared with B, with or without perturbation. There was no significant difference in MI($q_r,q_f$,K) and MI($q_r,q_f$,D) with or without perturbations. Since participants did not have prior knowledge of M or B, some outliers could have been due to the subject taking some time to determine how to successfully reach the target. Overall, these results strongly support \textbf{H1} and \textbf{H2}.

Finally, we investigated the perceived controllability of M and B (\textbf{H3}) among subjects; results for Q1-Q3 are summarized Figure~\ref{fig:metrics}-B. Subjects indicated that M provided a better match between motor intent and cursor motion, resulting in more timely execution of motor commands, and more precise control than B; results were more pronounced with perturbations. For Q4, six out of eight subjects gave a correct description of the perturbation field when using M while two subjects were unsure; with B, five out of eight subjects could not correctly describe the location of the force field and the others were unsure. For Q5, all the subjects had the same control strategy with B: adopt low muscle co-contraction and move the wrist until the joint limit is reached. With M, two subjects did not significantly increase muscle co-contraction, but the other six adapted joint impedance to counter perturbations. For Q6, all subjects agreed impedance modulation did not improve performance with B; two subjects stated that it resulted in the worst perceived controllability. With M, on the other hand, six out of eight subjects indicated that impedance adaptation helped counter perturbations; two subjects were unsure. These results support \textbf{H3} and correspond to the quantitative results described above (for \textbf{H1} and \textbf{H2}).

\subsection{Results for amputee}
Since only one amputee participated in the experiments, the values of performance measures obtained over the 40 trials per session are reported in Figure~\ref{fig:metrics}-A in blue. We observed that M provided a higher SR than B, with or without perturbation: $87.50\%$ and $65\%$ for M and B without perturbations, and $80\%$ and $55\%$ with perturbations. 
While there was an overall decrease in TP for the amputee compared with the able-bodied subjects, M provided a significantly higher TP compared with B, with or without perturbations. Also, while there was no significant difference in TR for the able-bodied participants, for the amputee M provided a significantly shorter TR than B with perturbations. M also provided a significantly higher PE than B, with or without perturbations. Notably, PE was comparable to or higher than with able-bodied participants. We observed no significant difference in Energy between M and B, whereas NM was significantly higher with B than with M in the presence of perturbations. These observations support \textbf{H1}. 
In the context of \textbf{H2}, we observed that SPARC and MI between $\tau_f$ and $q_r$ were significantly better with M than with B, with or without perturbation. There was no significant difference in MI($q_r,q_f$,K) and MI($q_r,q_f$,D) for M compared with B, with or without perturbations. Overall, these results support \textbf{H1} and \textbf{H2}.

Finally, Figure~\ref{fig:metrics}-C summarizes the amputee's responses to Q1-Q3; we asked the subject to answer questions four times per session (i.e., after every 10 targets) in an attempt to obtain more reliable answers. Similarly to the responses from able-bodied participants, the amputee indicated that M provided better controllability than B, and correctly described the force field (Q4) with M. 
For Q5, the amputee's control strategy when using B changed from tensing up the muscles to trying to minimally co-activate the muscles "or the cursor would jump too far"; this was an example of the baseline incorrectly assigning an increase in activation to a change in position. When using M, the amputee focused on co-contracting the muscles of the forearm when needed. For Q6, the subject was unsure if impedance modulation by muscle co-activation improved the performance with B since the cursor would sometimes oscillate unexpectedly; with M, however, she indicated three times that stiffening the muscles helped counter the perturbations, and mentioned that it once led to some overshoot. Overall, these results support \textbf{H3} and match the results reported for \textbf{H1} and \textbf{H2}.

\section{Discussion and future work}
\label{sec:conclusions}
We described a novel framework toward impedance control of upper-limb prostheses. The framework incorporates muscle-tendon units (MTUs) to estimate the human's motor intent from sEMG signals in terms of kinematics and dynamics and uses this information to implement an impedance controller on the robot. Unlike prior work, we enable simultaneous control of joint kinematics, joint stiffness, and damping. Also, the direct use of the stiffness and damping estimated from MTUs to implement the user's intent in the controller, ensures that the MTUs' dynamics better match those of the robot, improves the framework's transparency, and provides insights about learning motor intent from sEMG signals.
Experimental results obtained by comparing our framework with a state-of-the-art baseline during online reaching tasks in free space, and with external perturbations, demonstrated a significant improvement in performance for able-bodied subjects and an amputee, particularly in the presence of perturbations. Also, our framework allowed the participants to adapt stiffness and damping to modulate the physical interaction between the robot and the environment, contributing to improved performance compared with the baseline.
Note that we neither measure the human joint impedance nor claim to learn stiffness and damping values that match the biological ones. Instead, our computational framework provides a coherent representation of the dynamics of the MTUs and that of the robot, leading to improved controllability. 

Our framework provides interesting insights and opens up multiple directions for further research. For example, the conditional mutual information between $q_{r}$ and $q_{f}$ given the stiffness or damping provides an initial indication of the contribution of joint stiffness and damping to system stability. We plan to explore this more thoroughly by considering force fields that require specific joint stiffness or damping modulation (e.g. posture maintenance; reaching tasks in velocity-dependent force fields). 
In addition, our current framework has only considered one DoF. We plan to expand the framework to additional DoFs at the wrist, which is crucial for daily life tasks but makes the selection of sEMG signals and parameters' optimization more challenging. Furthermore, we will expand our participant group, which was limited by the COVID-19 pandemic and the difficulty in finding subjects willing to make the desired time commitment.

\acknowledgments{This work was supported in part by a studentship from the \textit{Physical Sciences for Health} Centre for Doctoral Training funded by the UK Engineering and Physical Sciences Research Council (EP/L016346/1). The authors thank Dr.~Deren Barsakcioglu, Dr.~Moon Ki Jung, Irene Mendez Guerra and Milia Helena Hasbani for technical support with the use of EMGs}
\appendix 
\newpage
\section{Appendix}
This document describes the supplementary material. Specifically, we provide:
\begin{enumerate}
    
     \item[\textbf{1}] An example of trajectory tracking results obtained through offline evaluation of the proposed framework (M) and a state-of-the-art data-driven baseline (B) for an able-bodied subject; 
    
    \item[\textbf{2}] An example of offline training of muscle-tendon model parameters without including the impedance controller in the framework M. We also show an offline evaluation of M on the same data when the impedance controller is included in the framework;
    
    \item[\textbf{3}] The time evolution of the values of variables in muscle-tendon space and joint space during a successful trial and during a failed trial for the reaching task with perturbation (with an able-bodied subject). We also show an example of a successful trial for the amputee in the presence of perturbations;
    
    \item[\textbf{4}] The time evolution of joint space variables for the baseline during a failed trial (with perturbation), and two successful trials (one each with and without perturbation) for an able-bodied subject. We show an example of a successful trial for the amputee in the presence of perturbations;
    
    \item[\textbf{5}] Additional results analysing the correlation between mutual information MI($\tau_{f}$,$q_{r}$) and the time to reach TR (for an able-bodied subject); and 
    \item[\textbf{6}] Details on the amputee and her feedback regarding the use of prostheses.
\end{enumerate}

\begin{figure*}[!htb]
\centering
\noindent\makebox[\textwidth]{\includegraphics[width=0.6\paperwidth]{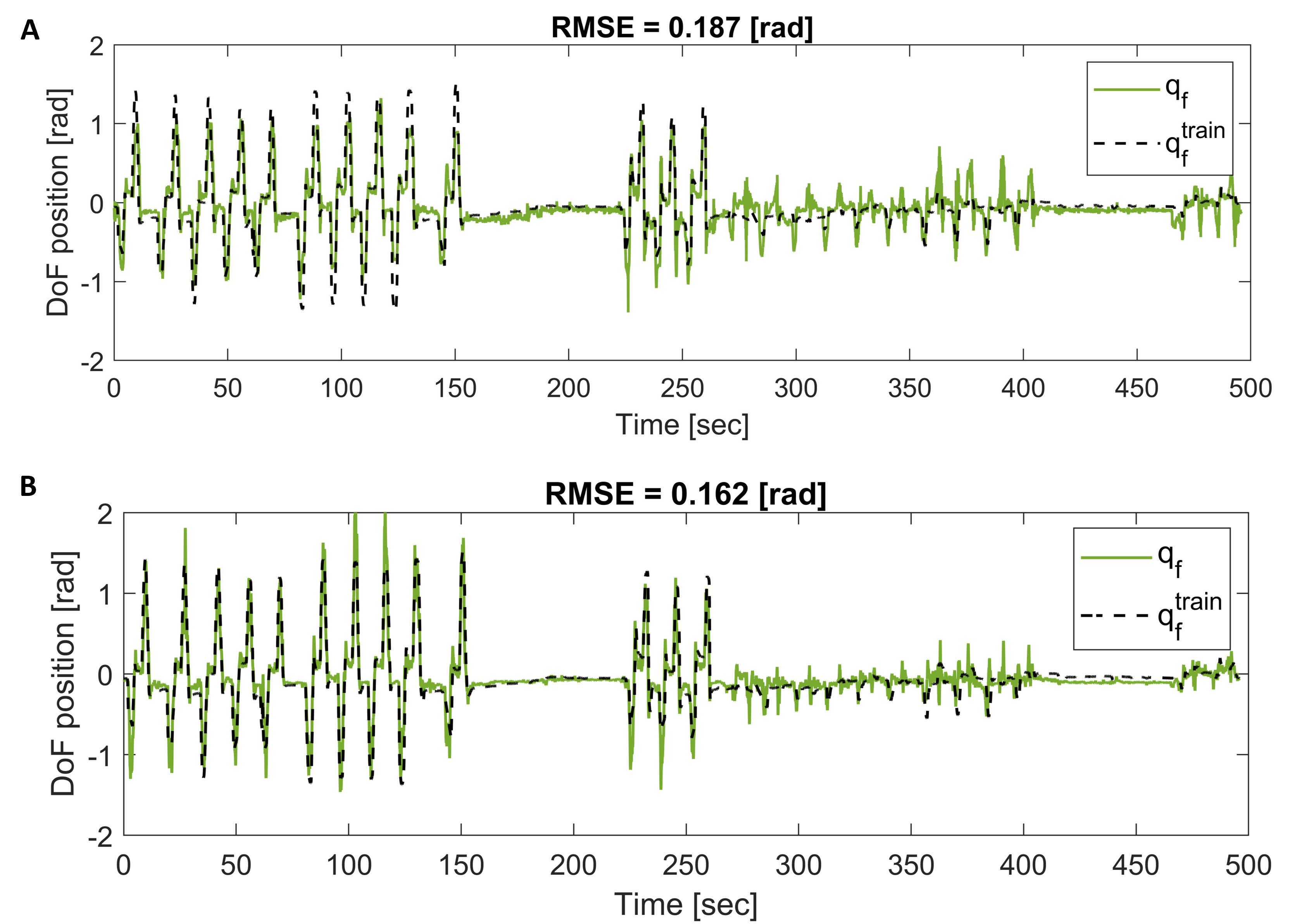}}
\caption{A) Trajectory tracking during offline evaluation of the \textbf{proposed framework (M)}. B) Trajectory tracking during off-line evaluation of the \textbf{baseline (B)}. The black dotted line is the ground truth position ($q_{f}^{train}$) of the flexion-extension DoF. The joint position trajectory ($q_{f}$) is obtained in each plot by evaluating the corresponding framework on the entire dataset.}
\label{fig:Mpred}
\end{figure*}

\subsection{Offline training results example}
Recall that the proposed framework includes a pair of muscle-tendon models to actuate a single Degree of Freedom (DoF) and a variable impedance controller for estimating human action intent from EMG signals and for adaptive, online execution of human intent in the presence of perturbations. The baseline, on the other hand, uses a neural network to directly maps EMG signals to a trajectory that is passed through a fixed (high stiffness) PD controller for execution. 

We first describe the trajectory tracking results obtained through offline training and evaluation of the proposed framework (M) and the baseline (B) for a representative subject. We record the EMG signals as a human participant executes a given motion pattern through the simulated interface; this motion primarily focuses on the flexion-extension DoF. We also record the corresponding ground truth pose using an external high-fidelity motion capture system. Part of the recorded data is used to learn the values of the framework's parameters; please see Figure 2 in the main paper. The trained framework (M or B) is then evaluated on the entire dataset, with the results summarized in Figure~\ref{fig:Mpred}-A,B. In these figures, the black dotted line represents the ground truth trajectory of flexion-extension DoF. We observe that B achieves a lower tracking error (RMSE), but overestimates the repetitions performed at high impedance, e.g., between 80-150 secs. 
During data collection, the subject performs ulnar-radial deviation motions of the wrist (DoF 1) from t = 275 seconds. We observe some indirect activation of DoF 2 (flexion-extension) during this interval because the two DoFs can not be completely decoupled. These indirect flexion-extension motions are overestimated by M, potentially because M trains separate muscle models based on specific EMG activations, whereas B jointly considers all eight EMG signals as inputs. However, B provides a more noisy estimate; it substantially over/under-estimates the trajectory in certain segments. 

\subsection{Importance of impedance controller}
\begin{figure}[!ht]
\includegraphics[width=1\columnwidth]{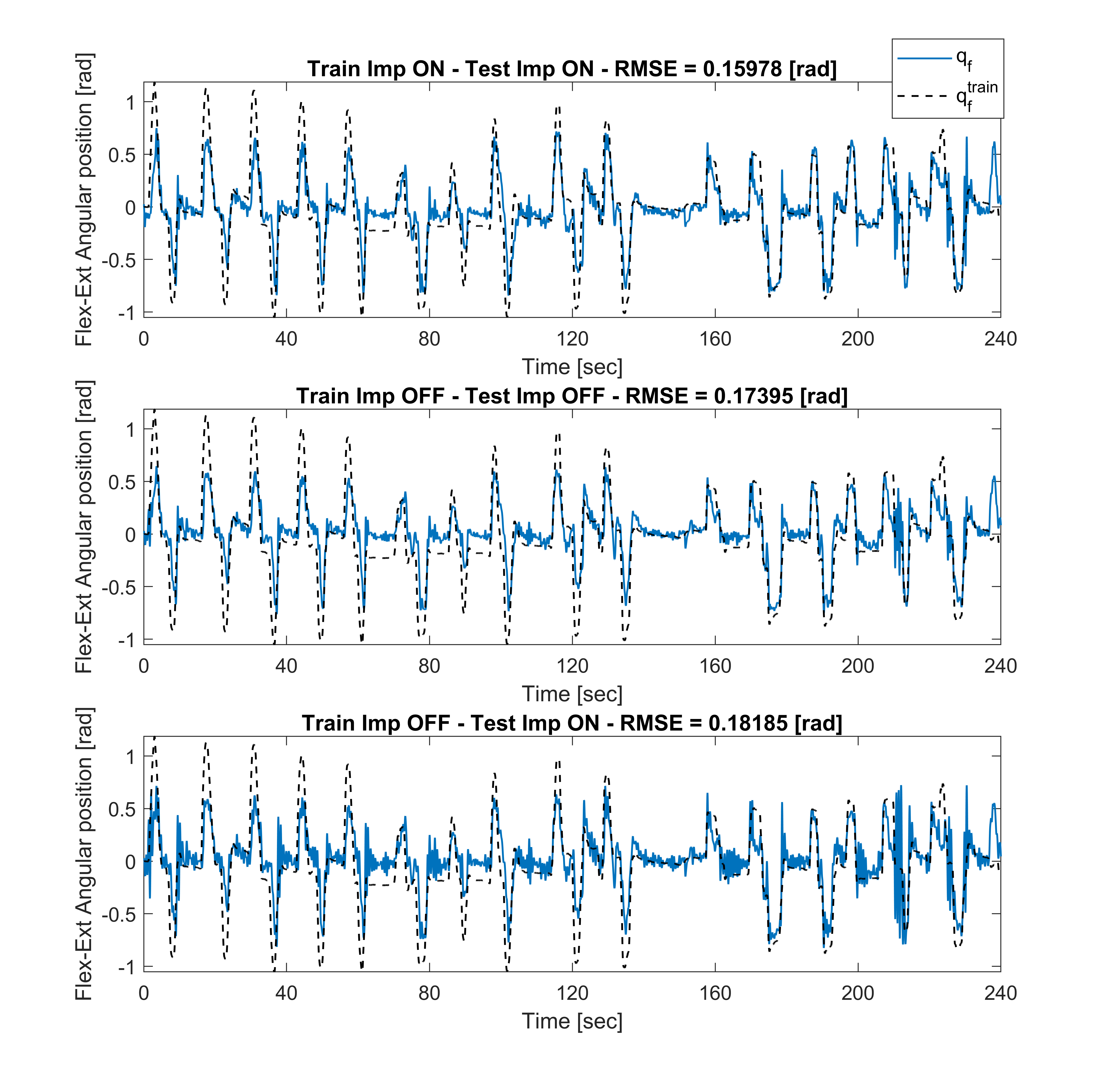}
\caption{Trajectory tracking during offline evaluation of the framework (M). The black dotted line is the ground truth position. The blue line is the predicted trajectory under the following conditions: (top) framework optimization and evaluation included the impedance controller:  $q_{f}$; (center) framework training and testing did not include the impedance controller: $q_{r}$; and (bottom) the optimization framework did not include the impedance controller, but the evaluation framework included the impedance controller: $q_{f}$.}
\label{fig:dynmatch}
\end{figure}

As discussed in Section 3.4 of the main paper, our optimization method uses the joint position trajectory of the robot's plant $q_{f}$ as an optimization signal. This choice is motivated by the need to use gains K and D in the position-based impedance controller and ensure that the dynamics of the muscle-tendon units (MTUs) support the implementation of a stable impedance controller. Existing methods (discussed in Section 2: Related work of the main paper) instead use the joint torque $\tau_{r}$ or $q_{r}$ as optimization signal.

We trained the MTUs using $q_{r}$ as optimization signal (60\% of collected data), i.e., the impedance controller and the robot's plant were not included in this training process. We then evaluated the trained MTUs on the entire dataset (for completeness) as part of the entire framework that includes the impedance controller and the robot's plant (Figure~\ref{fig:dynmatch}, third plot). We showed that the stiffness K and damping D cannot be used directly as gains in the impedance controller and that this leads to oscillatory behaviour and instabilities of the robot's plant. This explains why in related works the muscle-tendon stiffness were tuned to implement a position-based control on the robot. This solution allows stable control but does not support the key requirement of matching the MTUs' dynamics with the robot's dynamics. Figure~\ref{fig:dynmatch} shows the offline evaluation of the framework when the optimization method:
\begin{itemize}
    \item \textbf{included} the impedance controller, and the evaluation framework \textbf{included} the impedance controller.
    \item \textbf{did not include} the impedance controller, and the evaluation framework \textbf{did not include} the impedance controller.
    \item \textbf{did not include} the impedance controller, and the evaluation framework \textbf{included} the impedance controller.
\end{itemize}

\subsection{Evolution of state in proposed framework}
\begin{figure*}[!htb]
\centering
\noindent\makebox[\textwidth]{\includegraphics[width=0.7\paperwidth]{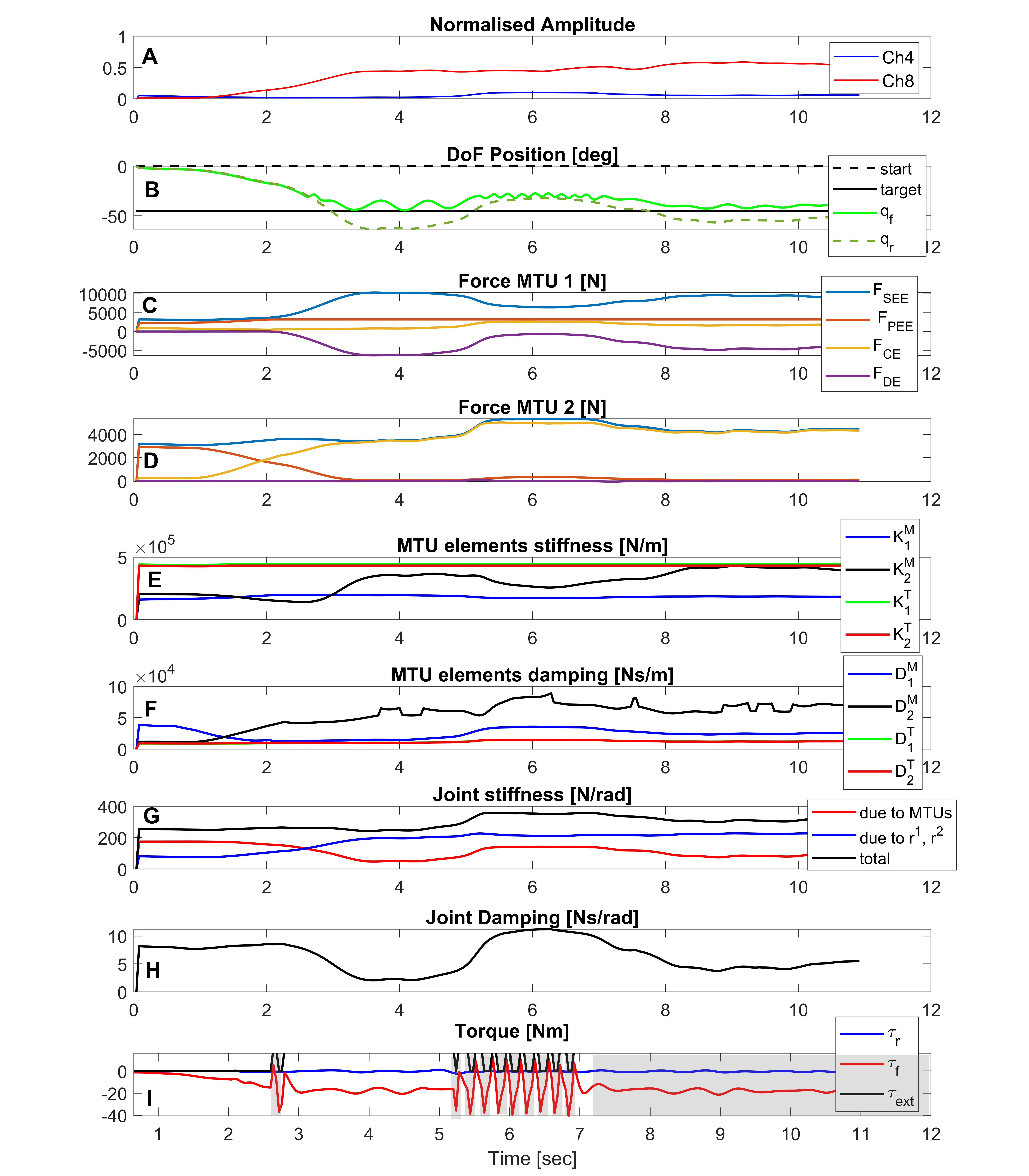}}
\caption{\textbf{Proposed framework (M).} Example of a successful trial. We plot the time evolution of the most relevant MTU  variables and controller/robot variables while an able-bodied subject performs a reaching task in the presence of a perturbation field. The system stabilises in less than two seconds.}
\label{fig:ex1M}
\end{figure*}
\begin{figure*}[!htb]
\centering
\noindent\makebox[\textwidth]{\includegraphics[width=0.8\paperwidth]{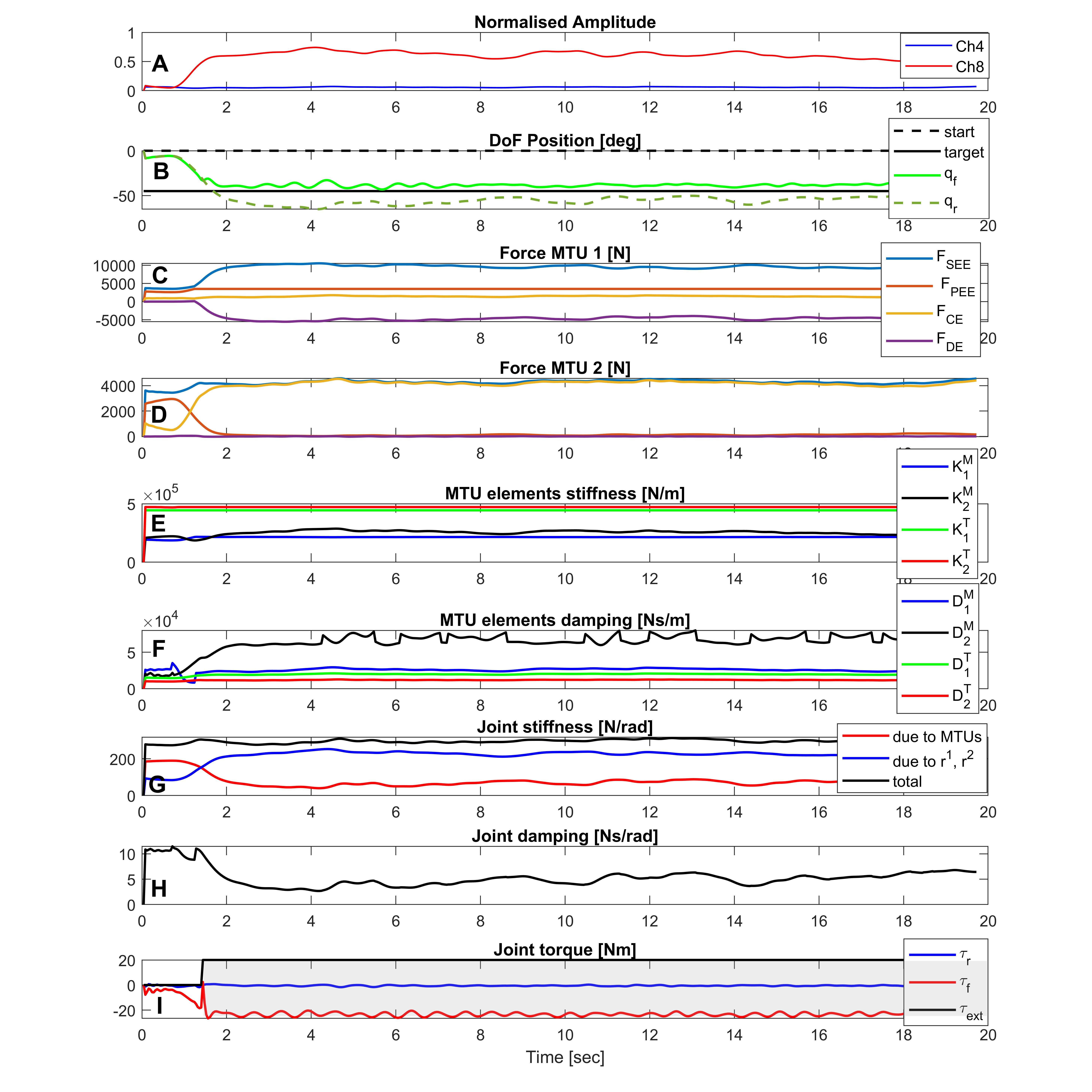}}
\caption{\textbf{Proposed method (M).} Example of failed trial. We plot the time evolution of the most relevant MTU variables and controller/robot variables while an able-bodied subject performs a reaching task in the presence of a perturbation field.}
\label{fig:ex2M}
\end{figure*}
We show plots corresponding to two illustrative examples of reaching tasks by an able-bodied subject in the presence of a perturbation field. In the first example in Figure~\ref{fig:ex1M}, the subject successfully reaches the target, whereas in the second example in Figure~\ref{fig:ex2M} the subject fails to reach the target. In each of these two figures, we plot:
\begin{itemize}
    \item[\textbf{A}] normalised input to the MTUs ($ch_{4},ch_{8}$) representative of the activity of the flexor and extensor muscles.

    \item[\textbf{B}] joint position $q_{r}$ (dotted green line) output by forward dynamics  and $q_{f}$ (continuous green line) output by the impedance controller, as the subject attempts to move $q_{f}$ from the initial position (black dotted line) to the target position (continuous black line). 

    \item[\textbf{C}] forces generated by the elements of $MTU_{1}$. 

    \item[\textbf{D}] forces generated by the elements of $MTU_{2}$. 

    \item[\textbf{E}] stiffness of $muscle_1$ ($K^{M}_{1}$) and $muscle_2$ ($K^{M}_{2}$), and the stiffness of $tendon_1$ ($K^{T}_{1}$) and $tendon_2$ ($K^{T}_{2}$). 

    \item[\textbf{F}] damping of $muscle_1$ ($K^{M}_{1}$) and $muscle_2$ ($K^{M}_{2}$), and the damping of $tendon_1$ ($K^{T}_{1}$) and $tendon_2$ ($K^{T}_{2}$). 

    \item[\textbf{G}] stiffness of the MTUs in the joint space (black signal), where the red signal is the stiffness contribution of the contraction dynamics of MTUs, and the blue signal is the stiffness contribution of the moment arms as a function of $q_{r}(t)$.

    \item[\textbf{H}] the damping of the MTUs in the joint space. 
    
    \item[\textbf{I}] the torque command provided by the impedance controller $\tau_{f}$ and the perturbation torque $\tau_{ext}$ due to the force field.
\end{itemize}
In the first example, the subject receives visual feedback on the effect of the perturbation and modifies their muscle activation in an attempt to negate the effect of the perturbation; the increase in co-activation in Figure~\ref{fig:ex1M}-A between 5-8 seconds corresponds to an increase in joint stiffness $K_{2}^{M}$ and damping $D_{2}^{M}$---see Figure~\ref{fig:ex1M}-G, H. In Figure~\ref{fig:ex1M}-B, we observe a discrepancy between $q_{r}$, unaffected by the perturbation field, and $q_{f}$, the output of the position-based impedance controller that "tracks" $q_{r}$ using the joint stiffness and damping based on the estimates provided by the MTUs. The error between the two trajectories decreases when the subject modulates the wrist impedance.

The time intervals during which the subject experiences the perturbation field is shaded gray in Figure~\ref{fig:ex1M}-I, i.e., $\tau_{ext}$ is not equal to zero. Figure~\ref{fig:ex1M}-D shows the forces generated by the elements of $MTU_{2}$, the muscle-tendon unit corresponding to flexion, i.e., contracting the muscle. We observe that the force generated by the active element and by the elastic component of the tendon, $F_{CE}$ and $F_{SEE}$ respectively, are mainly active. In the second example, on the other hand, only the (agonist) MTU $MTU_{2}$ is activated. We thus hypothesise that the \textit{low co-activation has an impact on task performance and the ability to counter the perturbation}, as the subject cannot generate the necessary torque to move through the perturbation field.

\begin{figure*}[!h]
\centering
\includegraphics[width=1\textwidth]{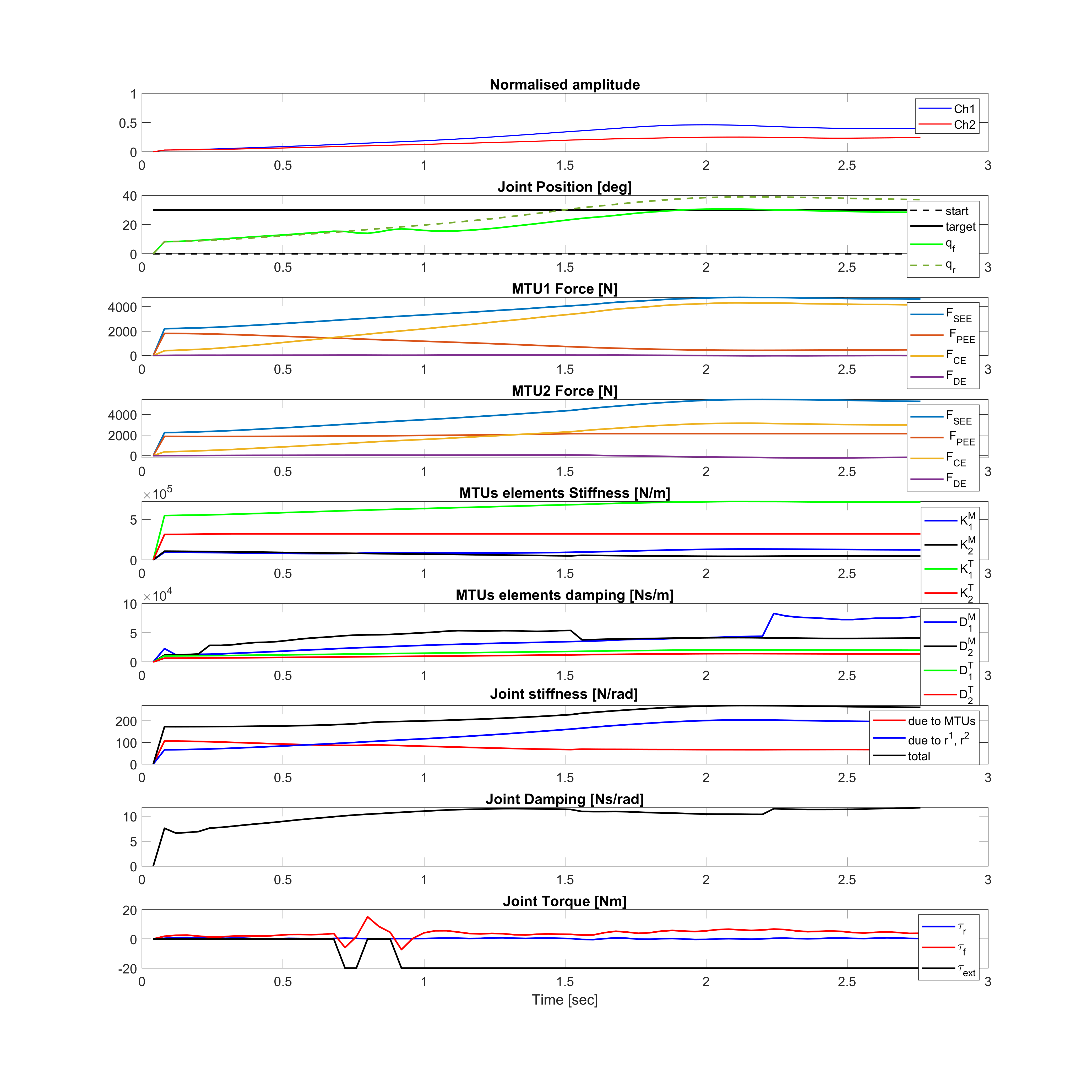}
\caption{{Proposed method (M).} Example of successful trial for the amputee. We plot the time evolution of the most relevant MTU variables and controller/robot variables while the subject performs a reaching task in the presence of a perturbation field.}
\label{fig:exMamputee}
\end{figure*}

\paragraph{Amputee trial example}
It can be observed in figure~\ref{fig:exMamputee} that in the case of a trial involving the amputee, the subject starts increasing the activation and coactivation from about 1.5 seconds into the trial. This results in an increase in stiffness and a change in damping that allows the subject to maintain the target position.

\subsection{Evolution of state in baseline}
\begin{figure*}[tb]
\centering
\noindent\makebox[\textwidth]{\includegraphics[width=0.5\paperwidth]{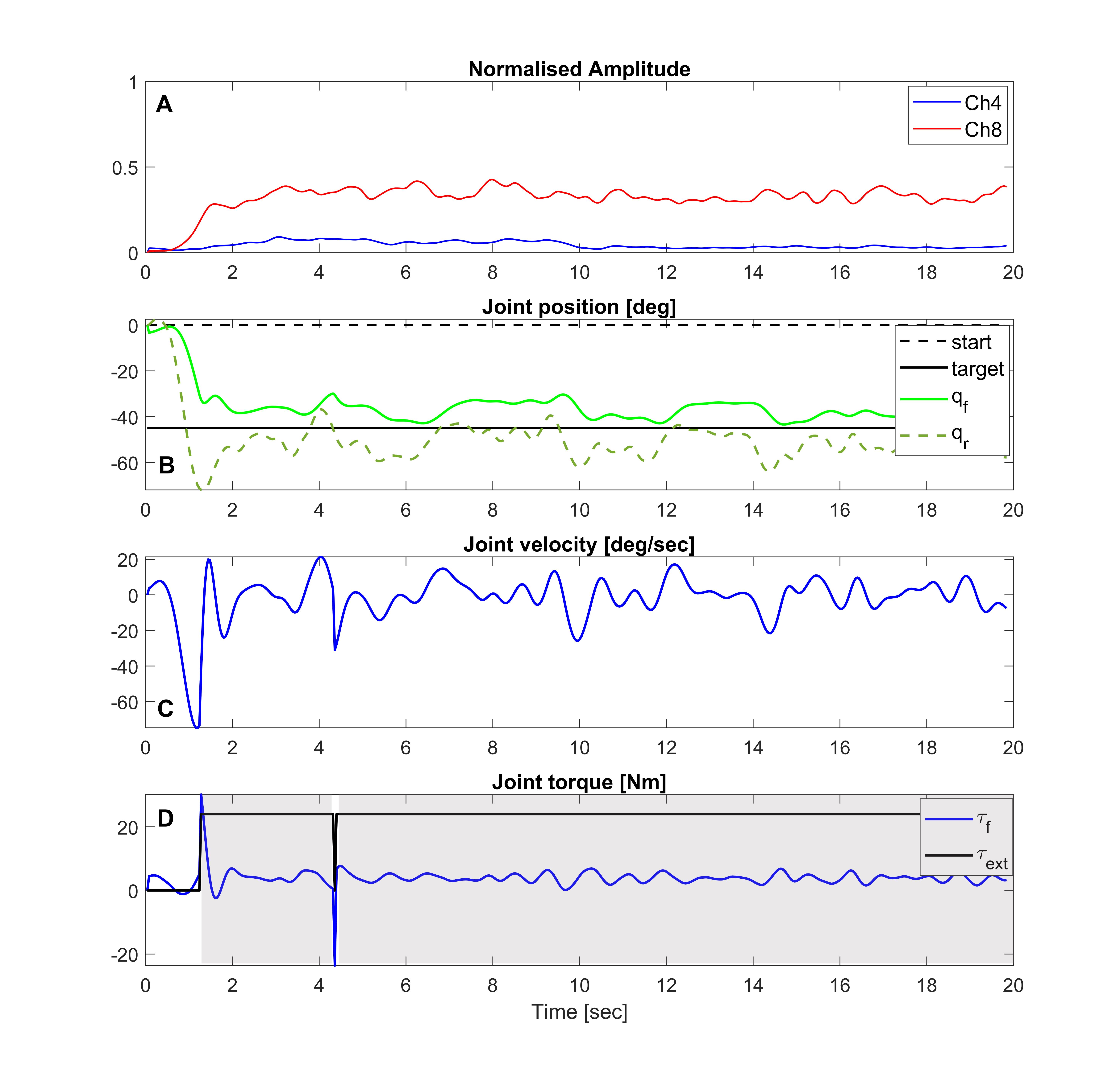}}
\caption{\textbf{Proposed method B.} Here we plot the time evolution of the controller/robot variables, while the subject performs a reaching task in the presence of a perturbation field. The subject fails to stabilize the system and passe through the force field.}
\label{fig:ex1B}
\end{figure*}

\begin{figure*}[!h]
\centering
\noindent\makebox[\textwidth]{\includegraphics[width=0.5\paperwidth]{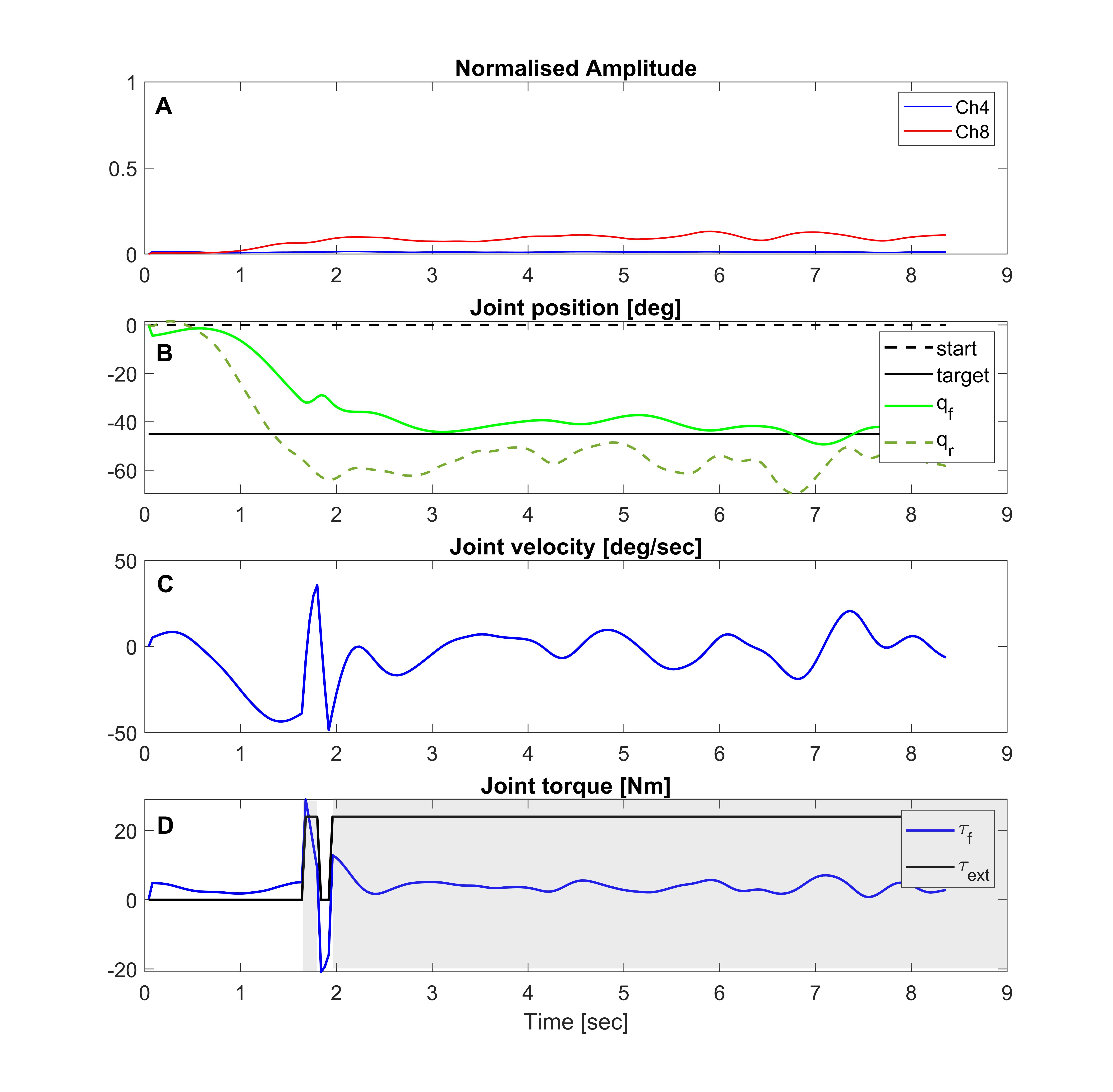}}
\caption{\textbf{Baseline B.} Time evolution of the controller/robot variables while the subject performs a reaching task in the presence of a perturbation field using baseline. Example of successful trial where the subject is able to counter the perturbation field.}
\label{fig:ex3B}
\end{figure*}

\begin{figure*}[!h]
\centering
\noindent\makebox[\textwidth]{\includegraphics[width=0.5\paperwidth]{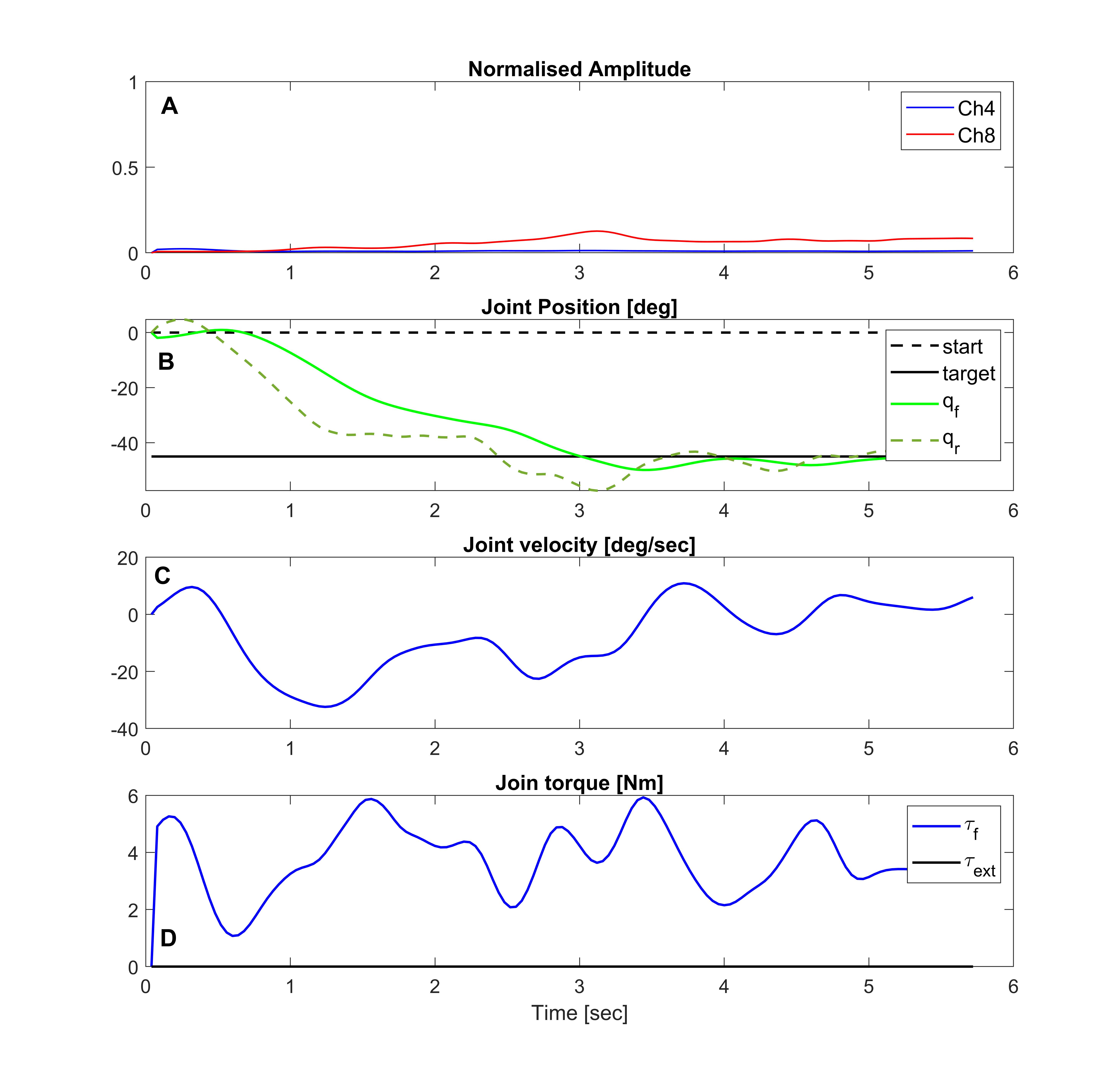}}
\caption{\textbf{Baseline B.} Time evolution of the controller/robot variables while the subject performs a reaching task in the absence of a perturbation field using baseline.}
\label{fig:ex2B}
\end{figure*}

We show plots corresponding to three illustrative examples of reaching tasks with the baseline framework: (i) Figure~\ref{fig:ex1B} shows a failed trial with perturbation; (ii) Figure~\ref{fig:ex3B} shows a successful trial in which the subject counters the effect of the perturbation field; and (iii) Figure~\ref{fig:ex2B} summarizes a successful trial in the absence of perturbations. In each figure, we plot the following signals:
\begin{itemize}
    \item[\textbf{A}] normalised EMG channels ($ch_{4},ch_{8}$) that represent the activity of the flexor and extensor muscles. Recall that the baseline considers all eight EMG signals from the myoband as input.
    
    \item[\textbf{B}] joint position $q_{r}$ (dotted green line) output by the neural network, $q_{f}$; (continuous green line) output by the PD controller when tracking $q_{r}$, as the subject attempts to move $q_{f}$ from the initial position (black dotted line) to the target position (continuous black line).
    
    \item[\textbf{C}] joint velocity.
     
    \item[\textbf{D}] torque output of PD controller ($\tau_{f}$) and the perturbation torque due to the force field ($\tau_{ext}$).
\end{itemize}
We observe that although the magnitude of $ch_{8}$ in Figure~\ref{fig:ex1B}-A is approximately five times the magnitude of $ch_{8}$ in Figure~\ref{fig:ex3B}-A, the reference joint position $q_{r}$ output by the neural network estimator is (on average) very similar. The difference is in the smoothness of the estimated trajectory; it is less smooth in Figure~\ref{fig:ex1B}-B, potentially due to the higher co-activation of the muscles. This may be the reason why the first trial resulted in a failure whereas the second one was successful. This experimental result, and our (potential) explanation of how co-contraction of muscles results in successful trials, reflect the feedback provided by the human subjects regarding their strategy to counter perturbations when using B (see the last paragraph of Sections 5.1 and 5.2 of the main paper).


\begin{figure*}[!h]
\centering
\noindent\makebox[\textwidth]{\includegraphics[width=0.75\paperwidth]{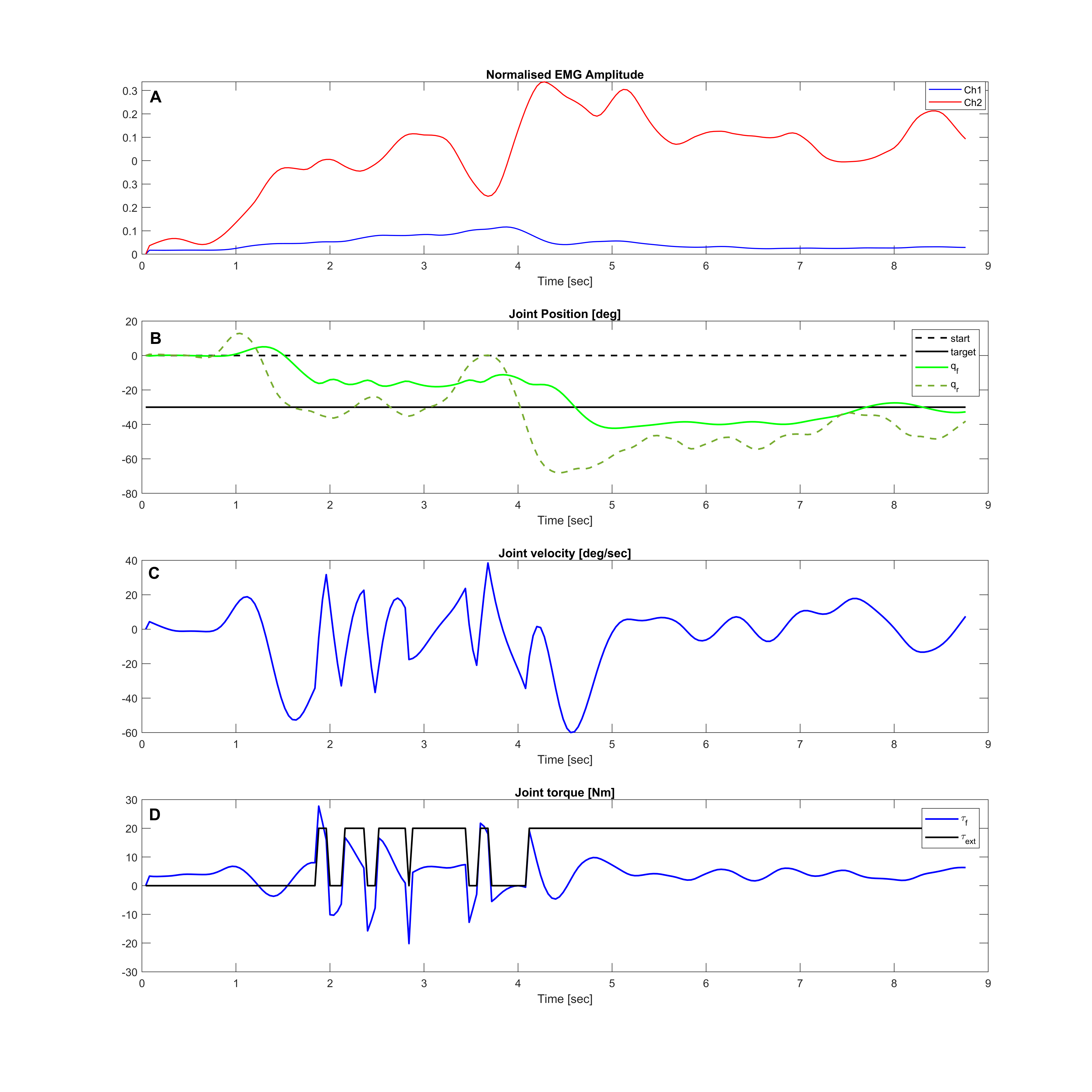}}
\caption{\textbf{Baseline B.} Time evolution of the controller/robot variables while the amputee performs a reaching task in the presence of a perturbation field with the baseline.}
\label{fig:exBamputee}
\end{figure*}

\paragraph{Amputee trial example}
It can be observed in Figure~\ref{fig:exBamputee} that the amputee attempts to increase the muscle activation to counter the force field, however, this increment in activation is mapped to a sudden change in position at $\approx$ 4.5 seconds. The amputee then decreases the muscle coactivation and manages to maintain the target position.

\clearpage
\subsection{Additional experimental results}
\begin{figure*}[!ht]
\centering
\noindent\makebox[\textwidth]{\includegraphics[width=0.4\paperwidth]{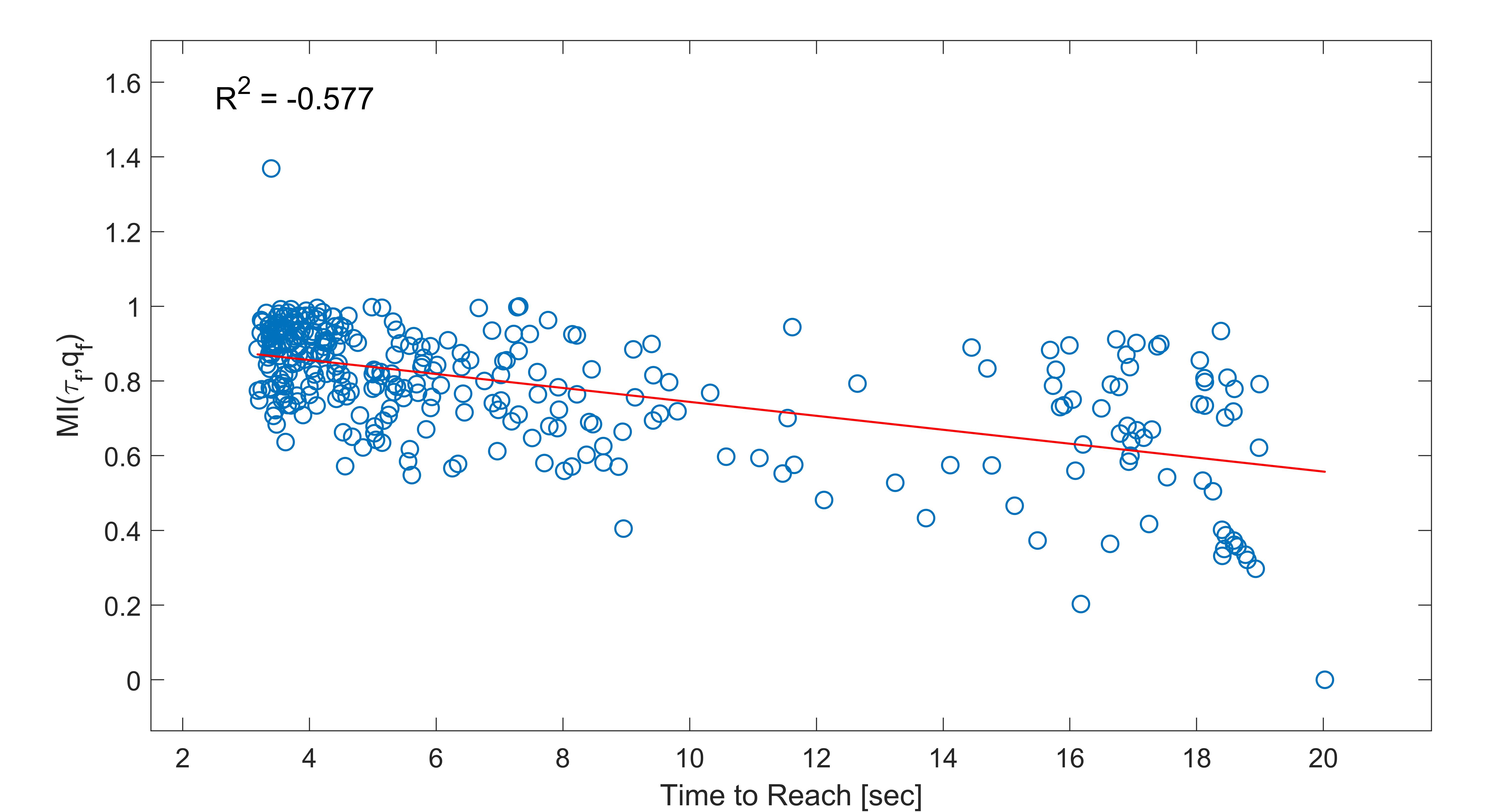}}
\caption{\textbf{Proposed framework M.} Scatter plot of the mutual information between $\tau_{f}$ (from impedance controller) and $q_{f}$, and the time to reach measure.}
\label{fig:corr_tau_M}
\end{figure*}

\begin{figure*}[!ht]
\centering
\noindent\makebox[\textwidth]{\includegraphics[width=0.4\paperwidth]{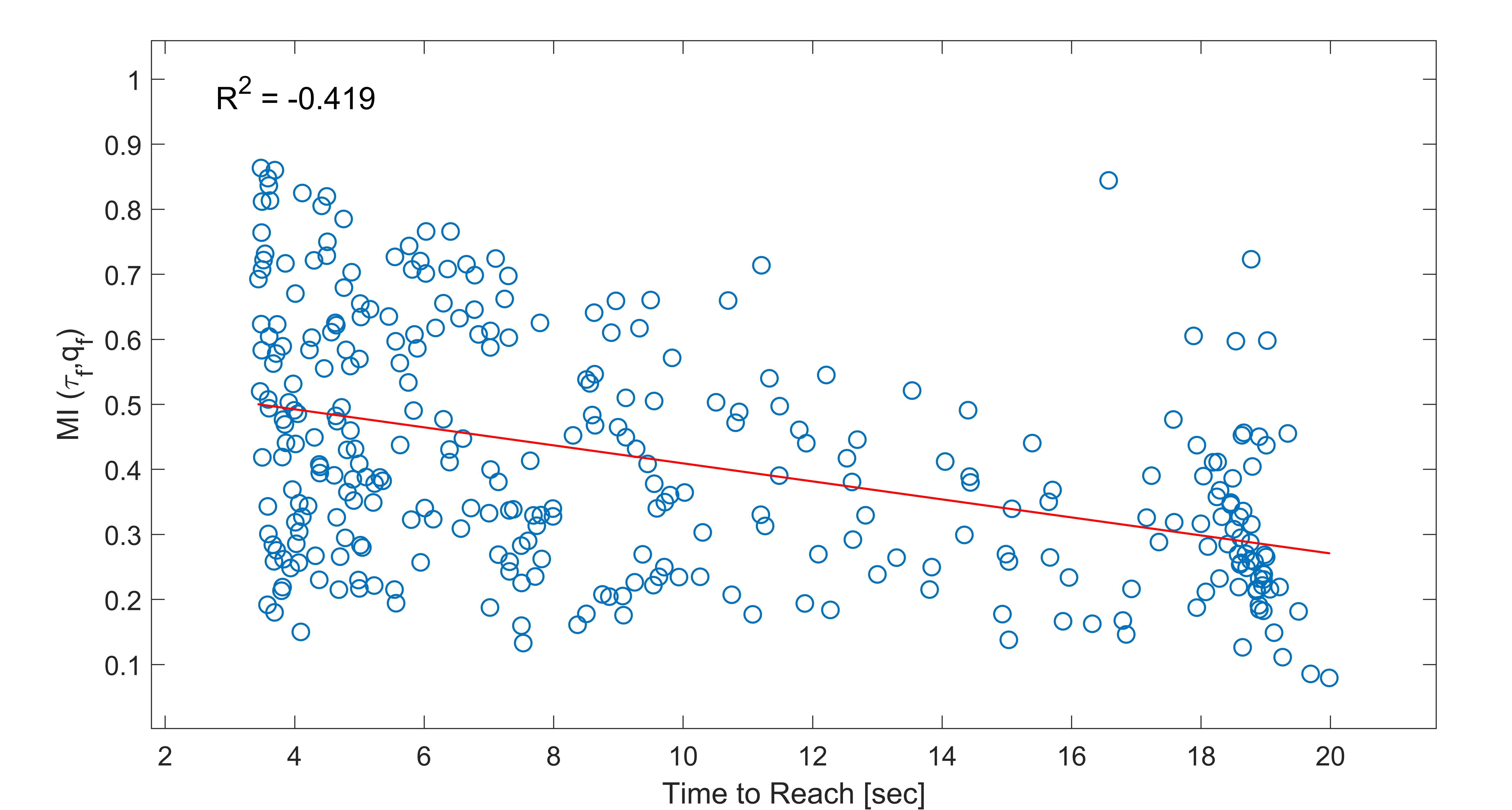}}
\caption{\textbf{Baseline B.} Scatter plot of the mutual information between $\tau_{f}$ (from PD controller) and $q_{f}$, and the time to reach measure.}
\label{fig:corr_tau_B}
\end{figure*}

Finally, we explored the relationship between the mutual information (between $\tau_{f}$ and $q_{f}$) and the Time to Reach (TR) measure, for both the proposed framework M and baseline B, as shown in Figure~\ref{fig:corr_tau_M} and Figure~\ref{fig:corr_tau_B} respectively. We notice that the experimental trials with a low "time to reach" have high mutual information for M, while there is no such correlation for B. For more information about the performance measures, please see Section 4.4 in main paper.

\subsection{Details of amputee and feedback on prosthesis use}
The participant's neurofibromatosis led to a transradial amputation of the left (not dominant) arm in 2016. Since then she has tried to use a prosthesis provided by the national sanitary system, but she decided to abandon the use because the functionality expectations were not enough compared to the psychological/physical overload required and discomfort. 
The participant stated that because of the amputation and disease she feels the remaining muscles of the forearm are constantly tight and tense. This might make it harder for her to perceive when she has low-medium-high stiffness.

On the other hand, please look at the feedback provided by the amputee after using our framework (Section 5.2 in the main paper). This indicates that our framework provides better controllability than the baseline, which is also what was reported by the able-bodied participants (Section 5.1). This is a promising result to be explored further in future work.
\newpage
\bibliography{sample-base}  

\begin{thebibliography}{38}
\providecommand{\natexlab}[1]{#1}
\providecommand{\url}[1]{\texttt{#1}}
\expandafter\ifx\csname urlstyle\endcsname\relax
  \providecommand{\doi}[1]{doi: #1}\else
  \providecommand{\doi}{doi: \begingroup \urlstyle{rm}\Url}\fi

\bibitem[Hogan(1984)]{hogan1984adaptive}
N.~Hogan.
\newblock Adaptive control of mechanical impedance by coactivation of
  antagonist muscles.
\newblock \emph{IEEE Transactions on automatic control}, 29\penalty0
  (8):\penalty0 681--690, 1984.

\bibitem[Hahne et~al.(2017)Hahne, Markovic, and Farina]{hahne2017user}
J.~M. Hahne, M.~Markovic, and D.~Farina.
\newblock User adaptation in myoelectric man-machine interfaces.
\newblock \emph{Scientific reports}, 7\penalty0 (1):\penalty0 1--10, 2017.

\bibitem[Muceli and Farina(2012)]{regress2}
S.~Muceli and D.~Farina.
\newblock {Simultaneous and proportional estimation of hand kinematics from EMG
  during mirrored movements at multiple degrees-of-freedom}.
\newblock \emph{IEEE Transactions on Neural Systems and Rehabilitation
  Engineering}, 20\penalty0 (3):\penalty0 371--378, 2012.

\bibitem[Jiang et~al.(2013)Jiang, Vujaklija, Rehbaum, Graimann, and
  Farina]{jiang2013accurate}
N.~Jiang, I.~Vujaklija, H.~Rehbaum, B.~Graimann, and D.~Farina.
\newblock Is accurate mapping of emg signals on kinematics needed for precise
  online myoelectric control?
\newblock \emph{IEEE Transactions on Neural Systems and Rehabilitation
  Engineering}, 22\penalty0 (3):\penalty0 549--558, 2013.

\bibitem[Capsi-Morales et~al.(2020)Capsi-Morales, Piazza, Catalano, Bicchi, and
  Grioli]{capsi2020exploring}
P.~Capsi-Morales, C.~Piazza, M.~G. Catalano, A.~Bicchi, and G.~Grioli.
\newblock Exploring stiffness modulation in prosthetic hands and its perceived
  function in manipulation and social interaction.
\newblock \emph{Frontiers in Neurorobotics}, page~33, 2020.

\bibitem[Hocaoglu and Patoglu(2022)]{hocaoglu2022semg}
E.~Hocaoglu and V.~Patoglu.
\newblock {sEMG-based natural control interface for a variable stiffness
  transradial hand prosthesis}.
\newblock \emph{Frontiers in Neurorobotics}, 16, 2022.

\bibitem[Karavas et~al.(2015)Karavas, Ajoudani, Tsagarakis, Saglia, Bicchi, and
  Caldwell]{karavas2015tele}
N.~Karavas, A.~Ajoudani, N.~Tsagarakis, J.~Saglia, A.~Bicchi, and D.~Caldwell.
\newblock Tele-impedance based assistive control for a compliant knee
  exoskeleton.
\newblock \emph{Robotics and Autonomous Systems}, 73:\penalty0 78--90, 2015.

\bibitem[Niu et~al.(2021)Niu, Luo, Chou, Liu, Hao, and
  Lan]{niu2021neuromorphic}
C.~M. Niu, Q.~Luo, C.-h. Chou, J.~Liu, M.~Hao, and N.~Lan.
\newblock Neuromorphic model of reflex for realtime human-like compliant
  control of prosthetic hand.
\newblock \emph{Annals of Biomedical Engineering}, 49\penalty0 (2):\penalty0
  673--688, 2021.

\bibitem[Osu et~al.(2002)Osu, Franklin, Kato, Gomi, Domen, Yoshioka, and
  Kawato]{osu2002short}
R.~Osu, D.~W. Franklin, H.~Kato, H.~Gomi, K.~Domen, T.~Yoshioka, and M.~Kawato.
\newblock Short-and long-term changes in joint co-contraction associated with
  motor learning as revealed from surface emg.
\newblock \emph{Journal of neurophysiology}, 88\penalty0 (2):\penalty0
  991--1004, 2002.

\bibitem[Hogan(1990)]{hogan1990mechanical}
N.~Hogan.
\newblock Mechanical impedance of single-and multi-articular systems.
\newblock In \emph{Multiple muscle systems}, pages 149--164. Springer, 1990.

\bibitem[Flash and Mussa-Ivaldi(1990)]{flash1990human}
T.~Flash and F.~Mussa-Ivaldi.
\newblock Human arm stiffness characteristics during the maintenance of
  posture.
\newblock \emph{Experimental brain research}, 82\penalty0 (2):\penalty0
  315--326, 1990.

\bibitem[Perreault et~al.(2001)Perreault, Kirsch, and
  Crago]{perreault2001effects}
E.~J. Perreault, R.~F. Kirsch, and P.~E. Crago.
\newblock Effects of voluntary force generation on the elastic components of
  endpoint stiffness.
\newblock \emph{Experimental brain research}, 141\penalty0 (3):\penalty0
  312--323, 2001.

\bibitem[Hogan(1985{\natexlab{a}})]{hogan1985mechanics}
N.~Hogan.
\newblock The mechanics of multi-joint posture and movement control.
\newblock \emph{Biological cybernetics}, 52\penalty0 (5):\penalty0 315--331,
  1985{\natexlab{a}}.

\bibitem[Hogan(1985{\natexlab{b}})]{hogan1985impedance}
N.~Hogan.
\newblock Impedance control: An approach to manipulation: Part
  ii—implementation.
\newblock 1985{\natexlab{b}}.

\bibitem[Furui et~al.(2019)Furui, Eto, Nakagaki, Shimada, Nakamura, Masuda,
  Chin, and Tsuji]{furui2019myoelectric}
A.~Furui, S.~Eto, K.~Nakagaki, K.~Shimada, G.~Nakamura, A.~Masuda, T.~Chin, and
  T.~Tsuji.
\newblock A myoelectric prosthetic hand with muscle synergy--based motion
  determination and impedance model--based biomimetic control.
\newblock \emph{Science Robotics}, 4\penalty0 (31):\penalty0 eaaw6339, 2019.

\bibitem[Tsuji et~al.(2010)Tsuji, Shima, Bu, and Fukuda]{tsuji2010biomimetic}
T.~Tsuji, K.~Shima, N.~Bu, and O.~Fukuda.
\newblock Biomimetic impedance control of an em-based robotic hand.
\newblock \emph{Robot Manipulators: Trends and Development}, page 213, 2010.

\bibitem[Hahne et~al.(2014)Hahne, Biessmann, Jiang, Rehbaum, Farina, Meinecke,
  M{\"u}ller, and Parra]{regress4}
J.~M. Hahne, F.~Biessmann, N.~Jiang, H.~Rehbaum, D.~Farina, F.~Meinecke, K.-R.
  M{\"u}ller, and L.~Parra.
\newblock Linear and nonlinear regression techniques for simultaneous and
  proportional myoelectric control.
\newblock \emph{IEEE Transactions on Neural Systems and Rehabilitation
  Engineering}, 22\penalty0 (2):\penalty0 269--279, 2014.

\bibitem[Smith et~al.(2015)Smith, Kuiken, and Hargrove]{regress5}
L.~H. Smith, T.~A. Kuiken, and L.~J. Hargrove.
\newblock Evaluation of linear regression simultaneous myoelectric control
  using intramuscular emg.
\newblock \emph{IEEE Transactions on Biomedical Engineering}, 63\penalty0
  (4):\penalty0 737--746, 2015.

\bibitem[Yeung et~al.(2022)Yeung, Guerra, Barner-Rasmussen, Siponen, Farina,
  and Vujaklija]{yeung2022co}
D.~Yeung, I.~M. Guerra, I.~Barner-Rasmussen, E.~Siponen, D.~Farina, and
  I.~Vujaklija.
\newblock Co-adaptive control of bionic limbs via unsupervised adaptation of
  muscle synergies.
\newblock \emph{IEEE Transactions on Biomedical Engineering}, 2022.

\bibitem[Krasoulis et~al.(2015)Krasoulis, Vijayakumar, and
  Nazarpour]{krasoulis2015evaluation}
A.~Krasoulis, S.~Vijayakumar, and K.~Nazarpour.
\newblock Evaluation of regression methods for the continuous decoding of
  finger movement from surface emg and accelerometry.
\newblock In \emph{2015 7th International IEEE/EMBS Conference on Neural
  Engineering (NER)}, pages 631--634. IEEE, 2015.

\bibitem[Xia et~al.(2018)Xia, Hu, and Peng]{ann1}
P.~Xia, J.~Hu, and Y.~Peng.
\newblock Emg-based estimation of limb movement using deep learning with
  recurrent convolutional neural networks.
\newblock \emph{Artificial organs}, 42\penalty0 (5):\penalty0 E67--E77, 2018.

\bibitem[Vujaklija et~al.(2018)Vujaklija, Shalchyan, Kamavuako, Jiang, Marateb,
  and Farina]{vujaklija2018online}
I.~Vujaklija, V.~Shalchyan, E.~N. Kamavuako, N.~Jiang, H.~R. Marateb, and
  D.~Farina.
\newblock Online mapping of emg signals into kinematics by autoencoding.
\newblock \emph{Journal of neuroengineering and rehabilitation}, 15\penalty0
  (1):\penalty0 1--9, 2018.

\bibitem[{Rohmer} et~al.(2013){Rohmer}, {Singh}, and {Freese}]{6696520}
E.~{Rohmer}, S.~P.~N. {Singh}, and M.~{Freese}.
\newblock V-rep: A versatile and scalable robot simulation framework.
\newblock In \emph{2013 IEEE/RSJ International Conference on Intelligent Robots
  and Systems}, pages 1321--1326, 2013.
\newblock \doi{10.1109/IROS.2013.6696520}.

\bibitem[Mat(2018)]{MATLAB:R2018b_u8}
\emph{{MATLAB version 9.5.0.1586782 (R2018b) Update 8}}.
\newblock The Mathworks, Inc., Natick, Massachusetts, 2018.

\bibitem[Hill(1938)]{hill1938heat}
A.~V. Hill.
\newblock The heat of shortening and the dynamic constants of muscle.
\newblock \emph{Proceedings of the Royal Society of London. Series B-Biological
  Sciences}, 126\penalty0 (843):\penalty0 136--195, 1938.

\bibitem[G{\"u}nther et~al.(2007)G{\"u}nther, Schmitt, and
  Wank]{gunther2007high}
M.~G{\"u}nther, S.~Schmitt, and V.~Wank.
\newblock High-frequency oscillations as a consequence of neglected serial
  damping in hill-type muscle models.
\newblock \emph{Biological cybernetics}, 97\penalty0 (1):\penalty0 63--79,
  2007.

\bibitem[Buchanan et~al.(2004)Buchanan, Lloyd, Manal, and
  Besier]{buchanan2004neuromusculoskeletal}
T.~S. Buchanan, D.~G. Lloyd, K.~Manal, and T.~F. Besier.
\newblock Neuromusculoskeletal modeling: estimation of muscle forces and joint
  moments and movements from measurements of neural command.
\newblock \emph{Journal of applied biomechanics}, 20\penalty0 (4):\penalty0
  367, 2004.

\bibitem[Sartori et~al.(2012)Sartori, Reggiani, Farina, and
  Lloyd]{sartori2012emg}
M.~Sartori, M.~Reggiani, D.~Farina, and D.~G. Lloyd.
\newblock Emg-driven forward-dynamic estimation of muscle force and joint
  moment about multiple degrees of freedom in the human lower extremity.
\newblock \emph{PloS one}, 7\penalty0 (12):\penalty0 e52618, 2012.

\bibitem[Scovil and Ronsky(2006)]{scovil2006sensitivity}
C.~Y. Scovil and J.~L. Ronsky.
\newblock Sensitivity of a hill-based muscle model to perturbations in model
  parameters.
\newblock \emph{Journal of biomechanics}, 39\penalty0 (11):\penalty0
  2055--2063, 2006.

\bibitem[Winters(1990)]{winters1990hill}
J.~M. Winters.
\newblock Hill-based muscle models: a systems engineering perspective.
\newblock In \emph{Multiple muscle systems}, pages 69--93. Springer, 1990.

\bibitem[Bennett1 et~al.(1986)Bennett1, Ker1, Imery, and
  Alexander1]{bennett11986mechanical}
M.~Bennett1, R.~Ker1, N.~J. Imery, and R.~M. Alexander1.
\newblock Mechanical properties of various mammalian tendons.
\newblock \emph{Journal of Zoology}, 209\penalty0 (4):\penalty0 537--548, 1986.

\bibitem[Rack and Ross(1984)]{rack1984tendon}
P.~Rack and H.~Ross.
\newblock The tendon of flexor pollicis longus: its effects on the muscular
  control of force and position at the human thumb.
\newblock \emph{The Journal of physiology}, 351\penalty0 (1):\penalty0 99--110,
  1984.

\bibitem[Lloyd and Besier(2003)]{lloyd2003emg}
D.~G. Lloyd and T.~F. Besier.
\newblock An emg-driven musculoskeletal model to estimate muscle forces and
  knee joint moments in vivo.
\newblock \emph{Journal of biomechanics}, 36\penalty0 (6):\penalty0 765--776,
  2003.

\bibitem[Van~Laarhoven and Aarts(1987)]{van1987simulated}
P.~J. Van~Laarhoven and E.~H. Aarts.
\newblock Simulated annealing.
\newblock In \emph{Simulated annealing: Theory and applications}, pages 7--15.
  Springer, 1987.

\bibitem[Kronander and Billard(2013)]{kronander2013learning}
K.~Kronander and A.~Billard.
\newblock Learning compliant manipulation through kinesthetic and tactile
  human-robot interaction.
\newblock \emph{IEEE transactions on haptics}, 7\penalty0 (3):\penalty0
  367--380, 2013.

\bibitem[Williams and Kirsch(2008)]{williams2008evaluation}
M.~R. Williams and R.~F. Kirsch.
\newblock Evaluation of head orientation and neck muscle emg signals as command
  inputs to a human--computer interface for individuals with high tetraplegia.
\newblock \emph{IEEE Transactions on Neural Systems and Rehabilitation
  Engineering}, 16\penalty0 (5):\penalty0 485--496, 2008.

\bibitem[Balasubramanian et~al.(2015)Balasubramanian, Melendez-Calderon,
  Roby-Brami, and Burdet]{balasubramanian2015analysis}
S.~Balasubramanian, A.~Melendez-Calderon, A.~Roby-Brami, and E.~Burdet.
\newblock On the analysis of movement smoothness.
\newblock \emph{Journal of neuroengineering and rehabilitation}, 12\penalty0
  (1):\penalty0 1--11, 2015.

\bibitem[Bazzi and Sternad(2020)]{bazzi2020human}
S.~Bazzi and D.~Sternad.
\newblock Human control of complex objects: towards more dexterous robots.
\newblock \emph{Advanced Robotics}, 34\penalty0 (17):\penalty0 1137--1155,
  2020.

\end{thebibliography}
\end{document}